# MD-Manifold: A Medical-Distance-Based Representation Learning Approach for Medical Concept and Patient Representation

Shaodong Wang, Ph.D.
Iowa State University
0074 Black Engineering Building
Ames, IA USA 50011
Email: shaodong@iastate.edu

Qing Li, Ph.D.
Iowa State University
3031 Black Engineering Building
Ames, IA USA 50011
Email: qlijane@iastate.edu

Wenli Zhang, Ph.D.
Iowa State University
3332 Gerdin, 2167 Union Drive
Ames, IA, USA 50011
Email: wlzhang@iastate.edu

Please send comments to Wenli Zhang at wlzhang@iastate.edu.

# MD-Manifold: A Medical-Distance-Based Representation Learning Approach for Medical Concept and Patient Representation

**Abstract**

The rise of HealthIT has made a large volume of medical data available, with distinct characteristics, presenting researchers and data analysts with an opportunity as well as a challenge - to effectively represent medical concepts and patients for healthcare analytical applications. Representing medical concepts for healthcare analytical tasks requires incorporating medical domain knowledge and prior information from patient description data. Current methods, such as feature engineering and mapping medical concepts to standardized terminologies, have limitations in capturing the dynamic patterns from patient description data. Other embedding-based methods have difficulties in incorporating important medical domain knowledge and often require a large amount of training data, which may not be feasible for most healthcare systems. Our proposed framework, MD-Manifold, introduces a novel approach to medical concept and patient representation. It includes a new data augmentation approach, concept distance metric, and patient-patient network to incorporate crucial medical domain knowledge and prior data information. It then adapts manifold learning methods to generate medical concept-level representations that accurately reflect medical knowledge and patient-level representations that clearly identify heterogeneous patient cohorts. MD-Manifold also outperforms other state-of-the-art techniques in various downstream healthcare analytical tasks. Our work has significant implications in information systems research in representation learning, knowledge-driven machine learning, and using design science as middle-ground frameworks for downstream explorative and predictive analyses. Practically, MD-Manifold has the potential to create effective and generalizable representations of medical concepts and patients by incorporating medical domain knowledge and prior data information. It enables deeper insights into medical data and facilitates the development of new analytical applications for better healthcare outcomes.

## 1. Introduction

In the past decade, information systems (IS) scholars have been extremely active in exploring novel ideas for machine learning in the context of various business and societal applications (Padmanabhan et al., 2022). The performance of machine learning methods is heavily influenced by the choice of data representation (e.g., feature selection or embedding generation) on which they are applied (Bengio et al., 2013). Consequently, representation learning has become a field in itself and a rapidly growing direction in machine learning and data science (Bengio et al., 2013; Y. Li et al., 2019). Discovering efficient representations of high dimensional concepts has been a key challenge in a variety of IS problem domains, such as text mining (Z. Wang et al., 2020), knowledge management (J. Li et al., 2020), language modeling (Samtani et al., 2022), theory integration (Ludwig et al., 2020) and multi-model data learning (Zheng & Hu, 2020).

As widely acknowledged, good data representations are expected to express general prior information and disentangle different explanatory factors of the data (Bengio et al., 2013; Bojanowski et al., 2017; Mikolov et al., 2013; Pennington et al., 2014). The prior information and the explanatory factors may not be task specific but are likely to be useful for a learning algorithm to solve downstream tasks. In healthcare systems, various types of data exist, including structured data (e.g., patients' demographic information and diagnosis or procedure codes) and unstructured data (e.g., clinical text, medical images, and vital sign signals). Medical concepts, which can be present in both structured and unstructured data, including medical codes/terms, drug names, and billing codes that are widely used in patient care, health services billing, public health statistics, and health services research, are a unique feature of medical data. However, medical concepts currently pose a bottleneck in medical data representation. Effectively representing medical concepts is a non-trivial task due to three important characteristics of them (Chute, 2005). First, medical concepts, by their nature, are complex notions with high dimensionality. For example, the WHO's International Classification of Diseases (ICD) is a widely used disease classification system with more than 17,000 ICD-9[1] codes (i.e., medical concepts). Besides, medical concepts typically contain complicated prior information, including levels of details, connected attributes, and hierarchical structures determined by medical domain knowledge. For example, the ICD-9 system maps diseases to general categories. Major ICD-9 categories include a set of comparable medical conditions; patient descriptions with sets of similar medical concepts typically reflect similar health issues. Moreover, many medical data contain a wealth of underlying patterns that pose complex data representation problems. Recovering or disentangling such underlying patterns is beneficial for downstream healthcare analytical tasks. For example, the cooccurrences of medical concepts constitute patient networks, disease networks, or drug networks, which are frequently used to assess the likelihood of the simultaneous presence of diseases for disease understanding or drug repurposing (García del Valle et al., 2019).

---
[1] ICD Ninth Revision

Learning compact and effective representations of medical concepts from patient descriptions of diagnoses or procedures (refer to as patient record hereafter), such as electronic health records, claims data, and pharmacy records, has a wide range of applications in healthcare analytics, including medical information retrieval and medical code referencing (Bai et al., 2019; Chute, 2005). In addition, multiple medical concepts define a patient's single medical event in a patient record (e.g., a hospital visit). A sequence of medical events further consists of a patient's medical history. Learning effective patient representations from patients' medical histories with many medical concepts is challenging and extremely useful across healthcare analytic tasks such as cohort selection, patient summarization, and healthcare outcome prediction (E. Choi et al., 2018; Freitas et al., 2020; Tang et al., 2018).

Effective representations of medical concepts require incorporating both medical domain knowledge and prior information inherent in patient description data, which are essential for downstream healthcare analytical tasks. Combining these two sources of information can result in more meaningful and task-agnostic representations. However, current methods have significant limitations. (1) Feature engineering is a way to take advantage of medical domain knowledge. Using feature engineering, medical researchers derive summary measures to represent patients' medical conditions (Mehta et al., 2018) (e.g., comorbidity index which is a linear combination of multiple medical concepts). However, the expressive power, performance, and generalizability of such methods are constrained by their linearity and designed applications. (2) Leveraging medical domain knowledge and the hierarchical structure of medical concepts, many healthcare analytics studies employ a two-step process to represent medical concepts. The first step involves mapping medical concepts to standardized medical terminologies with fewer dimensions, such as mapping 5-digit ICD-9 codes to 3-digit ICD-9 categories or Clinical Classifications Software (CCS) codes. The second step involves using one-hot encoding to represent patients and the medical concepts they correspond to (Rasmy et al., 2020). While these two aforementioned methods are commonly used, they have limitations in capturing the underlying prior information embedded in patient description data, such as the dynamic patterns of medical concept occurrences, which could be used as discriminative features to enhance performance on downstream analytical tasks. (3) Many natural language processing tasks have benefited from representing words as low-dimensional vectors, known as embeddings (Bojanowski et al., 2017; Mikolov et al., 2013; Pennington et al., 2014). These algorithms have also been extended to the healthcare analytical domain, where medical concepts found in patient descriptions are represented as vectors to facilitate healthcare-exploratory research and predictive modeling (Freitas et al., 2020; Tang et al., 2018). However, the prerequisites for generating word embeddings differ significantly from generating representations for medical concepts or patients. First, these methods typically assume a sequential relationship between words and the surrounding text, while medical concepts generated from a single medical event normally have a co-occurring relationship (e.g., medical codes in EHR and claims data do not have time-stamp precision beyond a daily level of granularity). Second, these methods are not specifically designed for medical concept representation and do not take into account medical domain knowledge during the representation learning process. Third, the majority of these algorithms are deep learning-based and often require a large amount of training data, which is not always feasible for most healthcare systems (E. Choi et al., 2018).

To cope with the shortcomings of previous approaches, guided by the design science research principles (Gregor & Hevner, 2013; Hevner et al., 2004), we propose a new framework, Medical-Distance-manifold (MD-manifold). Our framework leverages domain knowledge of medical concepts (i.e., the hierarchical structures of medical concepts) and crucial underlying patterns in patient records (i.e., the co-occurrence properties of medical concepts) to generate effective representations for both medical concepts and patients. The medical concept-level representations have important implications for healthcare-exploratory research, medical concept information retrieval, and medical code referencing. Meanwhile, the patient-level representations are useful in patient cohort selection and patient summarization and can enhance various downstream healthcare analytical tasks, such as healthcare outcome prediction, transfer learning for rare patient cohorts, and multimodal medical data fusion.

  Our work has significant contributions to IS research. First, our framework addresses the challenges associated with medical concept and patient representations in healthcare analytics. It aligns with the problem-solving paradigm of design science research, which emphasizes building IT artifacts in the healthcare context (Meyer et al., 2014), and is consistent with machine learning studies in IS that focus on domain-specific work in business and social sciences (Padmanabhan et al., 2022). By addressing an important application domain with existing limitations, our framework constitutes a significant contribution to the IS knowledge base. Second, our proposed framework places a strong emphasis on knowledge-driven machine learning, achieved through the incorporation of medical domain knowledge in representation learning. Our design principles provide a foundational design theory (Gregor & Hevner, 2013) and highlight the potential for incorporating rich domain knowledge in other problem domains within IS, such as financial data analysis, knowledge management, and legal information analysis. Our design principle has the potential to facilitate the development of impactful design artifacts in these problem domains. Third, our study, situated within computational design science research, contributes to the use of design science as a mechanism for creating middle-ground frameworks (Yang et al., 2022). As unstructured, complex, and high-dimensional data become increasingly prevalent, it becomes challenging to extract valid information from such data for IS research. Our study proposes a solution by generating data representations that can be effectively utilized in a wide range of downstream explanatory or predictive tasks, ultimately enhancing the impact of IS research. In practical terms, it is highly desirable to make medical concept and patient representations less task-specific to broaden the scope and increase the ease of applicability of healthcare analytical applications. Effective medical concept and patient representations can enable the rapid development of new applications and, more importantly, contribute to the progress towards better healthcare outcomes.

## 2. Literature review

### 2.1 The importance and limitation of representation learning in healthcare analytics

Since the mid-2000s, a wealth of healthcare data has emerged due to the widespread adoption of health IT, creating significant research opportunities for IS researchers. In recent years, IS research has increasingly focused on designing and utilizing algorithms and analytics for healthcare data (Baird et al., 2018), including healthcare predictive analytics (Bardhan et al., 2014), patient trajectory studies (Xie et al., 2021), omni-channel user journey analysis (Abbasi et al., 2019), and health digital trace analysis (Zhang & Ram, 2020). Medical concept- and patient-representations are crucial and have displayed encouraging outcomes in above-mentioned healthcare studies, but there are still research limitations. Effective medical concept representation techniques should automatically extract representations from raw input data that capture essential prior information in the data in an informative and efficient way. These techniques should produce features that are useful for a wide range of downstream tasks, robust to noise and input variations, and generalizable to new data. Additionally, they should be scalable, computationally efficient, and capable of learning from complex datasets (Bengio et al., 2013).

  Considerable research has been done on developing medical concept and patient representations (Table 1). One approach is to use summary measures to represent patients, while another common practice is to map medical concepts to higher-level code hierarchies or standard terminologies for representation, then represent patients with dimensionality-reduced medical concepts. These approaches use medical knowledge to categorize medical concepts and represent patients with dimensionality-reduced medical concepts, which are then used for downstream healthcare analytic tasks. The motivation for these approaches is comparable to ours: *to utilize and include medical domain knowledge into the representation learning process*. However, these methods have obvious restrictions in (1) identifying and disentangling underlying explanatory factors from patient records, limiting their ability to extract and organize discriminative information for generating better representations, and (2) generalizing learned representations across different healthcare tasks and domains, as they are often tailored to specific applications.

Table 1: Summary of existing methods in representing medical concepts and identified research gaps

| Reference | Methods | Data/Knowledge Requirements | Design motivation | Limitations / Advantages |
|---|---|---|---|---|
| (Charlson et al., 1987; Elixhauser et al., 1998; Sacco Casamassima et al., 2014; Sessler et al., 2010; van Walraven et al., 2009) | Deriving summary measures: CCI, ECI, RSI, … | • Demands a significant amount of medical expertise and specialist involvement<br>• Labeled patient records are not required | Utilize and include medical domain knowledge into the representation learning process | • Limited generalizability<br>• Inferior downstream predictions results<br>• Unable to capture dynamic data patterns |
| (Deschepper et al., 2019; Melton et al., 2006; Min et al., 2019; Rasmy et al., 2020; H.-H. Wang et al., 2019; Williams et al., 2017) | Mapping to higher level code hierarchy or standard terminologies[(i)]: ICD-9, CCS, CUI, SNOMED, … | • Existing medical domain knowledge<br>• Labeled patient records are not required | | • Limited generalizability<br>• Ambiguity of optimal mapping sources<br>• Unable to capture dynamic data patterns |
| (Bai et al., 2019, Y. Choi et al., 2016; Freitas et al., 2020; Si et al., 2021; Tang et al., 2018) | General embedding method[(ii)]: FastText, GloVe, Word2Vec, … | • Normally are supervised seq-to-seq models and do not necessitate labeled patient records | Allow representation learning algorithms to learn from large amounts of unlabeled/labeled patient records | • Variable-sized input data for downstream prediction models<br>• Large training data size<br>• No effective way to incorporate domain knowledge |
| (E. Choi et al., 2016, 2017, 2018) | Supervised deep learning method[(iii)] | • Labeled patient records are required | | |
| MD-Manifold (ours) | Manifold learning | • Existing medical domain knowledge<br>• An unsupervised learning task, and labeled patient records are not required | Incorporate both medical domain knowledge and prior data information | • Include both medical domain knowledge and prior data information<br>• Optimize representations using matrix factorization techniques, therefore, less demanding on the training data |

Note: (i) Most of these methods employ a two-step process: step 1, mapping medical concepts to standardized medical terminologies, and step 2, using one-hot encoding to represent patients and the medical concepts they correspond to. (ii) Most of these methods begin by producing embeddings for medical concepts and then aggregate variable-sized medical concepts' embeddings for patient representation. (iii) Utilize a healthcare prediction task as a training goal and extract one of the layers of a deep learning model as the representation of medical concepts; therefore, labeled patient records are required.

The recent advances of representation learning in natural language processing (NLP) provide alternative approaches to medical concept and patient representations. These approaches construct embeddings for individual medical concepts using NLP or deep learning techniques and create patient representations by combining multiple embeddings of medical concepts from the same patient record (Table 1). Similar to our proposed framework, *the motivation of these studies is to allow representation learning algorithms to learn from large amounts of patient records*, which can be fine-tuned for specific downstream healthcare analytical tasks. Concatenating medical concept embeddings for patient representations has also improved the performance of healthcare predictive tasks. However, these methods are not primarily designed for medical concept- and patient-representation, and thus have limitations in incorporating essential medical knowledge contained in medical concept ontologies into the representation learning process, which limit the ability of the resulting embeddings to improve the performance of downstream healthcare analytical tasks. Many of these algorithms rely on deep learning and demand considerable amounts of training data, a resource that may not always be accessible for many healthcare analytical tasks.

Addressing these research limitations is critical for advancing representation learning of medical concepts in healthcare analytics and for realizing its full potential in improving healthcare outcomes.

## 2.2. Manifold learning for representative learning

Figure 1: High-dimensional manifold spaces comprising medical concepts and patient records

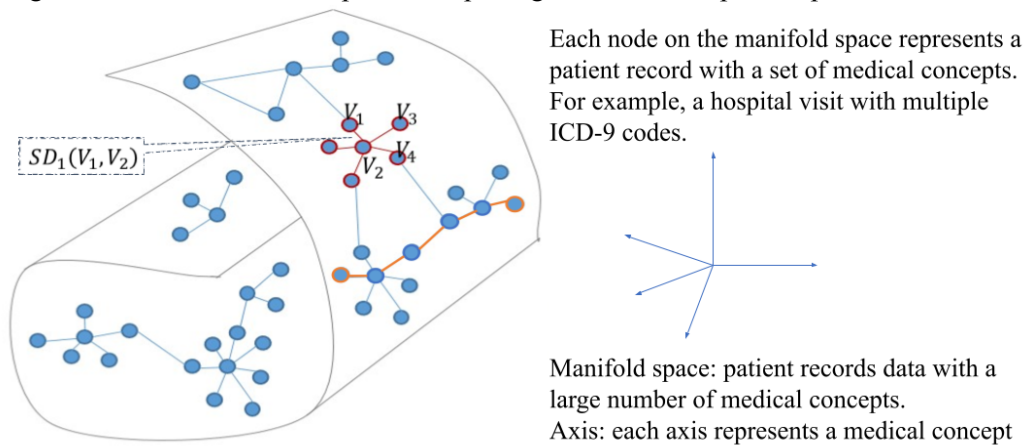

The local approach (e.g. LE) attempts to preserve the relationships between adjacent points as shown by the red circles, while the global approach (e.g. Isomap) attempts to preserve pairwise relationship as shown by the orange path.

Note: $SD_1(V_1, V_2)$ is the distance between patient record $V_1$ and $V_2$, see Formula (6).

Manifold learning is a promising way to overcome current limitations in medical concept and patient representation (Bengio et al., 2013; Talwalkar et al., 2013). Medical concepts in patient records form high-dimensional manifold spaces, where each data point represents a medical event in a patient record with various medical concepts (Figure 1). Manifold learning, as a non-linear representation learning approach, can capture the inherent structure of high-dimensional medical concepts (Silva & Tenenbaum, 2002). The manifold learning algorithms follow a similar pattern: first, they construct a nearest neighbor network to represent data points and second, provide a new representation for each point while preserving the network's internal structure (i.e., topology and geometry). Formally, given $n$ data points (e.g., $n$ patient records, each contains multiple medial concepts), $X = \{x_i\}_{i=1}^{n}$ and $x_i \in R^d$, the goal of manifold learning is to find corresponding outputs, $Y = \{y_i\}_{i=1}^{n}$ where $y_i \in R^k$, and $k \ll d$. Manifold learning algorithms generate low-dimensional space $Y$ to represent high-dimensional space $X$ while keeping data points' internal structure in $X$.

Manifold learning algorithms are categorized into two types: local and global approaches (Silva & Tenenbaum, 2002). Local approaches focus on mapping adjacent points from high-dimensional to nearby locations in low-dimensional space (Belkin & Niyogi, 2003; Donoho & Grimes, 2003; Roweis & Saul, 2000), while global approaches preserve the distance between adjacent points and map distant points to distant locations in low-dimensional space as well (Shaw & Jebara, 2009; Tenenbaum et al., 2000). Local approaches are computationally efficient but may not retain the global topography of the original data space. In contrast, global approaches tend to provide more reliable representations by preserving the global structure of the original manifold space. In our study, we compare Laplacian Eigenmap - a local method (Belkin & Niyogi, 2003) and Isomap - a global approach (Tenenbaum et al., 2000) to generate representations for patient records (Appendix A.1).

We aim to use manifold learning techniques to efficiently capture medical knowledge and prior information from patient records for healthcare analytical tasks. The construction of appropriate nearest neighbor networks is critical in manifold learning approaches (both local and global methods). These networks must incorporate relevant prior data information and medical domain knowledge to enable learned representations to capture and enhance essential medical knowledge and prior information from patient record data.

**2.3. Patient-patient network to incorporate medical knowledge and prior data information**

We propose the use of patient-patient networks as the nearest neighbor network for manifold learning to incorporate medical domain knowledge and important prior information in patient records. By doing so, we are motivated to include two types of critical information that are not task-specific but contain significant information that can enhance the performance of many downstream healthcare analytical tasks. (1) Medical domain knowledge from medical concepts' hierarchical structure: The hierarchical structure of medical concepts is an important property that reflects medical domain knowledge. For instance, the ICD-9 system categorizes diseases into generic categories, resulting in a well-organized hierarchy. Heart disease (i.e., medical concept; ICD-9 code "420-429"), for example, belongs to the circulatory system disease (i.e., medical concept; ICD-9 code "390-459") category. When generating representations for medical concepts and patients, it is crucial to consider the hierarchical structure of medical concepts as domain knowledge, so that the generated representations align well with medical knowledge and help downstream tasks reach better performance. (2) Prior information from patient description data: another important attribute of medical concepts is their co-occurrences in patient description data, which indicates a propensity for the simultaneous presence of two diseases in a patient. These co-occurrences also form patient-patient networks, commonly used to connect and evaluate comorbidities, making them crucial features for downstream healthcare analytical tasks.

Limited research has examined the benefits of considering both hierarchical structure and co-occurrence properties of medical concepts when generating representations of patients and concepts. Our research addresses this gap by considering both properties. We assume that similar medical concepts in patients' records indicate patients' similar health conditions. Additionally, patients' heterogeneities and homogeneities captured in patient records can enhance downstream healthcare analytical tasks by capturing patient-specific similarities and differences.

### 2.3.1. Patient-patient network

Patient-patient networks, a sub-research area of the human disease network, describe disease interconnections from an epidemiological perspective by constructing networks based on patients' similarities and differences (García del Valle et al., 2019). Various patient-patient networks have been developed, such as the patient-patient network used to identify type-2 diabetes (L. Li et al., 2015) and the patient-patient network constructed for precision medicine (Pai & Bader, 2018).

Patient-patient networks serve as optimal data structures for constructing patients' nearest neighbor networks, while preserving medical domain knowledge and essential prior information from patient records. The nodes in patient-patient networks typically represent patients (i.e., a node contains multiple medical concepts from a patient record), while the edges represent the similarities between patients (i.e., disease co-occurrence). *However, the existing patient-patient networks do not take into account the well-organized hierarchy of medical concepts as medical domain knowledge.* Therefore, we propose a new patient-patient network (i.e., nodes, edges, and edge weights) that embeds both the hierarchy of medical concepts and their co-occurrences. To build such a network, we need a novel distance metric for medical concepts and medical records. With an appropriate distance metric to calculate the distance between medical records, we can construct a nearest neighbor network for manifold learning algorithms, generate low-dimensional representations for medical concepts and patients, and employ the resulting representations in downstream healthcare analytical tasks.

### 2.3.2. Distance metrics of medical concepts

To calculate the distance between patient records with multiple medical concepts, two steps are involved: concept-level distance and record-level distance calculations (Jia et al., 2019). The former measures the distance between medical concepts, while the latter measures the distance between patient records based on the concept-level distance.

The objective of this study is to employ an appropriate concept-level distance metric that incorporates medical knowledge from the medical concept hierarchy and prior data information obtained from concepts' co-occurrences. Wu & Palmer (1994) and Y. Li et al. (2003) proposed the most similar concept-level distance metrics to our motivation, as they incorporated concept hierarchy for distance calculation. However, these approaches have limitations: their distance calculation solely depends on the hierarchical structure of medical concepts, ignoring their co-occurring frequencies in real-world data. Specifically, two distant concepts in the hierarchy may frequently co-occur in patient data in the real world, and a concept that appears more frequently than its siblings may have a closer relationship with its parent concept in the hierarchy. To address these limitations, we draw inspiration from previous work on semantic relatedness calculators (Patwardhan & Pedersen, 2006; Pedersen et al., 2007) and propose a novel concept-level distance metric. Our approach is both knowledge-driven, taking into account the hierarchical structure of medical concepts, and data-driven, considering the co-occurrences of medical concepts within the data. In addition, various record-level distance metrics exist for calculating the distance between sets of medical concepts based on the concept-level distance (Jia et al., 2019). Each distance metric holds significance, and their results may vary depending on the application. We explore their performance in Appendix A.2.

In summary, our goal is to preserve crucial medical domain knowledge and prior data information of medical concepts in patient records by utilizing appropriate medical concept distance metrics. To accomplish this, we introduce a novel medical concept distance metric and a patient-patient network. We integrate this new patient-patient network into manifold learning algorithms to create patient representations that can be utilized for downstream analysis.

## 3. Research design: Medical Distance-Manifold (MD-Manifold)

The MD-Manifold research design consists of two main components. First, we introduce a knowledge- and data-driven data augmentation method that preserves medical domain knowledge and essential prior data information in patient records, followed by generating medical concept representations based on the augmented data. Second, we propose a novel medical concept distance metric and patient-patient networks utilizing medical concept representations, followed by generating patient representations through adapting the patient-patient networks as nearest neighbor networks for manifold learning algorithms. The resulting medical concept and patient representations are suitable for a variety of downstream healthcare analytical tasks.

We denote a patient description dataset as $D$ with $n$ patient records $V_i$.

$$V_i = (M_1, M_2, ..., M_{h_i}) \; (i = 1, 2, ..., n) \tag{1}$$

A patient record $V_i$ contains a set of medical concepts $M_j$'s and describes a patient's health condition in a medical event, which can provide insight into a patient's current health condition. $M_j$ ($j = 1, 2, ..., h_i$) is a medical concept with the number of concepts to be $h_i \in [1, m]$, where $m$ is the maximum number of medical concepts for a $V_i$. Each $V_i$ contains different number of medical concepts, therefore $m = max_i(h_i)$. We then define the medical-concept hierarchy structure as a prefix tree $T$ (Fredkin, 1960) derived from the medical domain knowledge. A tree $T$ has a root node $N_{root}$, the internal nodes $N_{branch}$'s (i.e., branch nodes), and the terminal nodes $N_{leaf}$'s (i.e., leaf nodes, $N_{leaf}$'s are equivalent to $M_j$'s in $D$). The relations between the root, branch, and terminal nodes are represented as a set of linked nodes. Using different medical domain knowledge, we can construct different prefix trees, e.g., $T_{ICD9}$, $T_{CUI}$, and $T_{CCS}$ (details of the tree construction process and performance comparison are in Appendix A.3).

### 3.1. Medical concept representation

Effective representations of medical concepts have various healthcare analytic applications, such as medical information retrieval and medical code referencing. To achieve a representation that incorporates task-agnostic medical domain knowledge and prior information from patient records, we propose a novel data augmentation method considering the hierarchical structure and co-occurrence of medical concepts in dataset $D$. This process is knowledge- and data-driven. We then utilize random projection to reduce the dimensionality of the medical concept representation, preserving the key information while reducing the computational burden of downstream medical concept-level healthcare analytical tasks.

### 3.1.1. Knowledge-driven occurrence matrix construction

We first construct an occurrence matrix $O_{n \times N}$ for all $V$ in $D$, where $n$ is the number of $V$, and $N$ is the total number of medical concepts in the data. Denote each element of $O$ as $O_{ij}$, where $i$ is the index of the medical record, and $j$ is the index of the medical concept. First, for each $V_i$, we augment it by adding all the ancestors ($N_{branch}$'s) of its medical concepts, resulting $V_i^a$. Then, we set $O_{ij} = 1$ if the $j^{th}$ concept occurs in the augmented $V_i$, otherwise, $O_{ij} = 0$.

$$V_i^a = (M_1, ..., M_j, ..., M_{h_i}, N_{branch\_1}, ..., N_{branch\_j}, ..., N_{branch\_h_i}) \ (i = 1, 2, ..., n) \quad (2)$$

Figure 2: An example of $T_{ICD9}$ and an augmented $V_i^a$

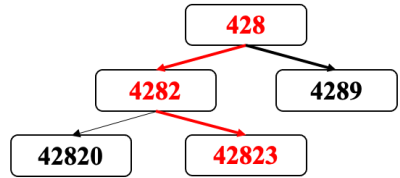

$V_i = \{42823, \times\times\times, \times\times\times, \times\times\times\}$

$V_i^a = \{42823, 4282, 428, \times\times\times, \times\times\times, \times\times\times\}$

Note: (1) The augmented $V_i$ retains the path in the prefix tree $T_{ICD9}$ from 42823 to 4282 and 428.

(2) xxx in $V_i$ represents the other medical concepts in the same record. The ICD-9 code 428 is not at the highest level in $T$. For simplicity, we do not show the parent concepts of 428.

Figure 2 presents an example of $V_i^a$. If $V_i$ contains an ICD-9 code 42823, the augmented $V_i^a$ contains ICD-9 codes 4282 and 428, which are the ancestors of the ICD-9 code 42823. The red line shows the path in the prefix tree $T_{ICD9}$ from 42823 to 4282 and 428. Figure 3 serves as an illustrative example. Suppose we have a dataset, $D$. The second column in Figure 3 (a) shows the medical concepts (e.g., ICD-9 codes) that belong to each record $V_i$, and the third column shows the corresponding frequencies. The first row indicates that there are ten records in the dataset that contain both medical concepts 4289 and 42823. We insert their ancestors into $V_i$ to obtain $V_i^a$ (Figure 3 (b)). Then we obtain the occurrence matrix $O$ in Figure 3 (c).

The purpose and advantage of constructing the occurrence matrix, $O$, are to retain medical domain knowledge in the tree $T$ for generating medical concept and patient representations later. This process is knowledge-driven because (1) it maintains path information in the prefix tree $T$, and (2) by adding all ancestors of a medical concept into the corresponding medical record $V_i$, it incorporates categorical disease information or higher-level medical concept ontologies into the augmented $V_i^a$, depending on how the medical domain knowledge tree $T$ is built (see Appendix A.3).

### 3.1.2. Data-driven co-occurrence matrix construction

In the next step, we construct a co-occurrence matrix $C_{N \times N}$ by computing the co-occurrences of medical concepts in the dataset $D$, where $N$ is the total number of medical concepts. The co-occurrence matrix is calculated using the occurrence matrix $O$, where $C = O^T O$. The resulting matrix $C$ is symmetric, and the non-diagonal elements represent the co-occurrences of medical concepts. Each row, $c_a$, of the matrix $C_{N \times N}$ (where $N$ is the total number of medical concepts) is a vector representation of a medical concept that contains the hierarchical structure of $T$ and the cooccurrences of medical concepts.

$$c_a = (c_{M_a M_1}, ..., c_{M_a M_j}, ..., c_{M_a M_N}), a \in [1, N] \tag{3}$$

The construction of the co-occurrence matrix $C$ is a data-driven process because its elements (i.e., co-occurring frequencies) are derived from the dataset $D$. The co-occurrence matrix $C$ has important implications in healthcare analytical tasks. First, the co-occurrence of diseases in a patient record $V_i$ is often referred to as comorbidity or multimorbidity in clinical practice, such as the co-occurrences of anxiety and depression and the co-occurrences of functional impairment and mortality (John et al., 2003). Second, comorbidity is often associated with linked diseases at the molecular level (Barabási et al., 2011), providing implicit information for downstream healthcare analytical tasks (Goh et al., 2007).

The use of augmented $V_i^a$ (section 3.1.1) in matrix $O$ confers two benefits to the construction of matrix $C$. First, the augmented $V_i^a$ contains disease categorical information or higher-level medical concept ontologies, resulting in the inclusion of co-occurring relationships between diseases and disease categories in the co-occurrence matrix $C$. Second, consider two medical concepts, $M_j$ and $M_{j'}$ (i.e., both are leaf nodes in $T$), which are rare in the dataset $D$ but have a frequent co-occurrence. This relationship can be challenging to capture due to the rarity of these concepts, and the potential variation in their co-occurrence with other medical concepts in $D$. However, it is easier to represent the co-occurring relationship between their parent nodes $N_{branch\_j}$ and $N_{branch\_j'}$ in the augmented $V_i^a$ using the co-occurrence matrix $C$. This is because $M_j$ and $M_{j'}$'s sibling concepts share the same parent nodes (sibling concepts indicate the nuanced difference in the medical knowledge), increasing the probability of $N_{branch\_j}$ and $N_{branch\_j'}$'s appearing in $V_i$. Such co-occurrence relationships are crucial for medical-concept distance calculations because the vector representations of $M_j$ and $M_{j'}$ should have a relatively short distance in the manifold feature space since they tend to co-occur. The co-occurring relationships are important medical domain knowledge that we strive to preserve when obtaining the medical concept and patient representations using patient records.

### 3.1.3. Medical concept representation using random projection matrix

$C_{a_{1 \times N}}$, where $N$ is the number medical concepts in dataset $D$, is an expressive vector representation of a medical concept that contains both medical domain knowledge and prior data information. However, its high dimensionality makes it impractical for downstream tasks. We address this issue by using a random projection matrix $R_{N \times k}$, where $k \ll N$ and components of $R_{N \times k}$ are drawn from the distribution $N\left(0, \frac{1}{w}\right)$, where $w$ is the number of components in $R$. Based on the Johnson-Lindenstrauss lemma, $R_{N \times k}$ maps $c_a$ into a lower-dimensional space, while retaining the pairwise distances between points (Bingham & Mannila, 2001; Dasgupta, 2013). We consider each row, $c_a'$, of $C_{N \times k}$ as the representation of each medical concept in dataset $D$. The use of random projection to reduce the dimensionality of $C_a$ while retaining its information enables $C_a$ to be more computationally efficient for downstream concept-level healthcare analytical tasks.

$$C'_{N\times k} = C_{N\times N} R_{N\times k}$$
$$c_a' = (c_{a1}, ..., c_{ak}), a \in [1, N]$$
(4)

Our approach to generating concept representations is both knowledge-driven and data-driven, making it not only useful for medical concepts but also for other representation learning tasks, such as representing Industrial Classification Codes in Finance and legal terms in legal research. By considering both the domain knowledge and co-occurrence information in the data, our method can provide better representations of these concepts for downstream analyses.

**3.2. Patient representation generation**

A patient's health conditions can be represented in a patient record $V_i$, which comprises multiple medical concepts. Using the augmented medical concept representation, $c_a$, that we learned in the previous step, we can create an effective representation of multiple medical concepts in $V_i$ to represent the patient. The process of learning effective patient representations from patient records that contain numerous medical concepts is challenging yet crucial for healthcare analytic tasks, including patient cohort selection, patient summarization, and healthcare outcome prediction.

To generate patient representations, we create a patient-patient network that serves as the input for manifold learning algorithms, as discussed in sections 2.2 and 2.3. Constructing the patient-patient network involves two steps. First, we calculate the distance between medical concepts. Second, we use the distance between medical concepts to calculate the distance between patient records, which represents the similarity of the patient's health condition.

Deriving suitable distances among medical concepts is crucial for incorporating medical domain knowledge and prior information in $D$. Nevertheless, the most state-of-the-art concept distance metric (Y. Li et al., 2003; Wu & Palmer, 1994) has limitations. As discussed in Section 2.3.1, high co-occurrence frequencies of medical concepts in the real-world patient description data indicate their close relationships. However, using existing metrics, the distance of two medical concepts is solely determined by their relative positions in the concept hierarchy $T$, which does not reflect their co-occurring frequencies in the real-world observational data. To overcome this limitation, we define a new group of medical-concept distance metric $CD_{new}$ that considers both the medical concepts' hierarchical structure and co-occurrences in the dataset $D$.

### 3.2.1 Medical concept distance calculation

We begin by calculating the distance between medical concepts using the co-occurrence matrix $C$. It is worth noting that our patient representation generation process is based on manifold learning, which inherently involves dimensionality reduction. Therefore, in this step, we utilize $c_a$, which provides the most comprehensive information from medical domain knowledge and prior data information, instead of $c_a'$ to construct the patient-patient network. We introduce a medical-concept distance metric that incorporates the augmented medical-concept representation in the manifold space. The distance metric is defined as follows:

$$
\begin{aligned}
CD_{new} &= distance(c_a, c_b) \\
CD_{new-Cosine} &= 1 - \frac{C_a \cdot C_b}{\sqrt{C_a \cdot C_a}\sqrt{C_b \cdot C_b}} \\
CD_{new-Manhattan} &= ||C_a - C_b||_1 \\
CD_{new-Euclidean} &= ||C_a - C_b||_2 \\
CD_{new-eHDN} &= 1 - \frac{C_{a,b}N - \Sigma C_a \Sigma C_b}{\sqrt{\Sigma C_a \Sigma C_b (N - \Sigma C_a)(N - \Sigma C_b)}}
\end{aligned}
\tag{5}
$$

Where $a$ and $b$ are two medical concepts, and $C_a$ and $C_b$ are row $a$ and row $b$ of the co-occurrence matrix $C$, respectively. The $distance(\cdot)$ functions can be defined in various ways. In Appendix A.2, we compare four distance formulas: $CD_{new-Cosine}$, $CD_{new-Manhattan}$, $CD_{new-Euclidean}$, and $CD_{new-eHDN}$ to determine which ones are better suited for healthcare analytical tasks.

Figure 3 (e) shows an example of $CD_{new-Cosine}$ given the co-occurrence in Figure 3 (d). Notice that the concept 42823 occurs more frequently than 42820 in Figure 3 (a). It is reasonable to believe that patients with an upper-level concept 4282 are more likely to have 42823 than 42820 as a specified disease, which indicates that the concept 4282 is more related to 42823 than 42820. By using our method $CD_{new-Cosine}$, as we expected, (42823, 4282) has a smaller distance than (42820, 4282) with $CD_{Cosine}(42823, 4282) = 0.0125$ and $CD_{Cosine}(42820, 4282) = 0.2463$. Moreover, due to the higher co-occurrence frequency of (4289, 42823) than (4289, 42820), $CD_{Cosine}(4289, 42823) = 0.0458$ is smaller than $CD_{Cosine}(4289, 42820) = 0.425$, in spite of the equal-distance relationship in the medical concept hierarchy $T_{ICD9}$, which overcomes the limitation of existing distance metric (Wu & Palmer, 1994) (i.e., without taking into account their co-occurring frequencies, the distance between two medical concepts is exclusively determined by their positions in $T$).

Figure 3: An illustrative example of medical concept distance calculation and patient-patient network construction

(a) Patient records with medical concepts

| Patient record $V_i$ | Medical concepts $M_j$'s | Frequency |
|---|---|---|
| $V_1$ | {4289, 42823} | 10 |
| $V_2$ | {4289} | 2 |
| $V_3$ | {42823} | 10 |
| $V_4$ | {42820} | 5 |

(b) Augmented patient records

| Augmented $V_i$ | $V_i^a$ | Frequency |
|---|---|---|
| $V_1^a$ | {4289, 42823, 4282, 428} | 10 |
| $V_2^a$ | {4289, 428} | 2 |
| $V_3^a$ | {42823, 4282, 428} | 10 |
| $V_4^a$ | {42820, 4282, 428} | 5 |

(c) The Knowledge-driven occurrence matrix, $O$

| 4289 | 42823 | 428 | 4282 | 42820 | Frequency |
|---|---|---|---|---|---|
| 1 | 1 | 1 | 1 | 0 | 10 |
| 1 | 0 | 1 | 0 | 0 | 2 |
| 0 | 1 | 1 | 1 | 0 | 10 |
| 0 | 0 | 1 | 1 | 1 | 5 |

(d) The data-driven co-occurrence matrix, $C$

|  | 4289 | 42823 | 428 | 4282 | 42820 |
|---|---|---|---|---|---|
| 4289 | 12 | 10 | 12 | 10 | 0 |
| 42823 | 10 | 20 | 20 | 20 | 0 |
| 428 | 12 | 20 | 27 | 25 | 5 |
| 4282 | 10 | 20 | 25 | 25 | 5 |

| 42820 | 0 | 0 | 5 | 5 | 5 |

(e) Medical concept distance calculation

e.g., $CD_{new-Cosine}(4289, 42823) = 1 - \frac{C_{4289} \cdot C_{42823}}{\sqrt{C_{4289} \cdot C_{4289}}\sqrt{C_{42823} \cdot C_{42823}}} = 0.0458$

|       | 4289   | 42823  | 428    | 4282   | 42820  |
|-------|--------|--------|--------|--------|--------|
| 4289  | 0      | 0.0458 | 0.0523 | 0.0652 | 0.4250 |
| 42823 | 0.0458 | 0      | 0.0133 | 0.0125 | 0.3595 |
| 428   | 0.0523 | 0.0133 | 0      | 0.0014 | 0.2495 |
| 4282  | 0.0652 | 0.0125 | 0.0014 | 0      | 0.2463 |
| 42820 | 0.4250 | 0.3595 | 0.2495 | 0.2463 | 0      |

(f) Patient record distance calculation for patient-patient network construction

e.g., $SD_1(V_1, V_2) = \frac{CD_{new-Cosine}(4289, 42823)}{|V_i|+|V_j|} = 0.0153$

### 3.2.2 Patient-patient network generation

In the second step, we construct a patient-patient network and prepare it for manifold learning algorithms. To construct a patient-patient network, we first measure distances among patient records (Formula (6)) using the medical concept-level distance (Formula (5)). Then we construct the patient-patient network by connecting similar medical records based on the patient record distance:

$$SD = distance(V_i, V_j)$$

$$SD_1 = \frac{1}{|V_i|+|V_j|} \left( \sum_{a \in V_i} \min_{b \in V_j} CD(a,b) + \sum_{b \in V_j} \min_{a \in V_i} CD(b,a) \right)$$

$$SD_2 = \frac{1}{|V_i \cup V_j|} \left( \sum_{a \in V_i \setminus V_j} \frac{1}{|V_j|} \sum_{b \in V_j} CD(a,b) + \sum_{b \in V_j \setminus V_i} \frac{1}{|V_i|} \sum_{a \in V_i} CD(b,a) \right) \quad (6)$$

$$SD_3 = \frac{1}{|V_i| \cdot |V_j|} \sum_{a \in V_i, b \in V_j} CD(a,b)$$

$$SD_4 = \frac{1}{|MWBM|} \sum_{(a,b) \in MWBM} CD(a,b)$$

Where each record, $V_i$ or $V_j$, comprises a set of medical concepts. We compare four widely used metrics for sets of medical concepts distance calculation (Jia et al., 2019). $SD_1$ and $SD_4$ are designed to capture the similarities of the most similar medical-concept pairs from two medical records. $SD_2$ does not include the overlapping medical concept but focuses on the difference between two medical records. $SD_3$ is widely used in clustering analysis in measuring the distance between each cluster (also known as "Average Linkage"). Because each metric has its merit in finding the distance between medical records, we compare them in the pilot experiments (Appendix A.2).

Then, given the distance, $SD$, between the patient records, we are able to construct the patient-patient network $G_{SD}$. Specifically, after we compute the distance for each pair of medical records in the dataset, we find $k$ neighbors with the shortest distance for each medical record. We construct the network by regarding each patient record, $V_i$, as a node and connecting each pair of neighbors as an edge.

### 3.2.3 Patient representation using Manifold learning

Next, the manifold learning algorithms take the constructed network, $G_{SD}$, as the input to generate the representations that preserve the topology of the original patient-patient network. Mathematically,

$$Y = ML(G_{SD}) \qquad (7)$$

Where $ML$ is the manifold learning algorithm, and $Y = \{Y_i\}_{i=1}^{n}$ is the patient representation result. Formally, we denote a patient record as $V_i$, and the corresponding representations as $Y_i$. We connect two vertices $V_i$ and $V_j$ with an edge $E_{ij}$ if $V_i$ is the k-nearest neighbor of $V_j$ or vice versa, determined by the distance $SD$. The resulting vertices and edges form the network $G_{SD}(V, E)$. Please note that different combinations of $CD$ and $SD$ can lead to different patient-patient networks. We compare the performance of different $G_{SD}$'s in the pilot experiments (Appendix A.2). Then we apply manifold learning algorithms to find low-dimensional representations of $V_i$'s (Equation 7). Laplacian Eigenmap minimizes the objective function, $\Phi(Y) = \sum_{i,j} ||Y_i - Y_j||$, which is the total distance between connected vertices (i.e. k-nearest neighbors of each other) in the low-dimensional space. Isomap solves the objective function $\Phi(Y) = \sum_{i \neq j}(d_{ij} - ||Y_i - Y_j||)^2$, where $d_{ij}$ is the shortest distance between two records $V_i$ and $V_j$ in the network $G_{SD}(V, E)$. Laplacian Eigenmap and Isomap have different strategies to optimize the representations. Laplacian Eigenmap preserves the relationships of close neighboring nodes, while the Isomap maintains the shortest distance between each pair of nodes. Such a difference explains why Laplacian Eigenmap is a local approach and Isomap is a global approach. Both Laplacian Eigenmap and Isomap have advantages and disadvantages, as discussed in Section 2.2. We examine and compare their performance for healthcare analytical tasks in the pilot experiments (Appendix A.1). The generated patient representation $Y_i$ are ready to be used as the input of downstream healthcare analytical tasks.

## 4. Evaluations and Results

Following the design science approach (Gregor & Hevner, 2013), we evaluate the operational utility of our proposed method in two ways. First, we assess the quality of the generated representations from two perspectives: (1) we evaluate the quality of the medical *concept-level representations* in terms of their alignment with existing medical domain knowledge, and (2) we examine *patient-level representations* to determine their ability to distinguish heterogeneous patient groups, which may benefit downstream healthcare analytical tasks. Second, we evaluate the applicability of the generated representations and their ability to enhance the performance of *downstream applications* by conducting three healthcare analytical tasks. (1) First, we compare the performance of patient-level presentations in healthcare outcome prediction with state-of-the-art methods. (2) Second, we test whether the generated patient representations can aid in a transfer learning task for predicting rare disease patient cohorts' healthcare outcome. (3) Finally, we examine how the generated patient representations can add value to other clinical data modalities, such as text data and demographic data, in a multimodal healthcare data fusion task. Additional experimental results, including the pilot study comparing model components and the details of the experimental settings, can be found in Appendix A and B, respectively.

For *concept-level representations* evaluation, we follow the benchmark methods used in previous research (Bai et al., 2019; E. Choi et al., 2016) and compare MD-Manifold with state-of-the-art medical concept representation techniques (Bai et al., 2019; E. Choi et al., 2016, 2016; Manning et al., 2008; Mikolov et al., 2013; Pennington et al., 2014) in finding the most similar ICD-9 codes with the CCS codes. For *patient-level representation* evaluation, we evaluate MD-Manifold against eight state-of-the-art baseline methods (Table 2), categorized into three types. (1) Feature engineering and summary measures: Charlson Comorbidity Index (CCI) (Sundararajan et al., 2004), Elixhauser Comorbidity Index (ECI) (Mehta et al., 2018), and Risk Stratification Index (RSI) (Verdecchia, 2003). They are commonly used comorbidity severity measures to predict mortality risk. (2) Code mappings followed by one-hot encoding: 4 or 5-digit ICD-9 codes are mapped to two standard medical terminologies, including 3-digit ICD-9 codes[2] and CCS codes[3] (Rasmy et al., 2020). (3) Embeddings of medical concepts: including three state-of-the-art studies. First, the element-wise sum of individual medical concepts' embeddings generated from Word2Vec, GloVe, and FastText (Tang et al., 2018). The second is Phe2Vec, which is a state-of-the-art unsupervised embedding method. It is based on the weighted sum of embeddings of medical concepts (Freitas et al., 2020). The last is Med2Vec, a supervised deep-learning method that uses a two-layer neural network to generate representations for medical concepts and records (E. Choi et al., 2016).

Table 2: Benchmarks in patient-level representation evaluation

| Category | Name and reference | Note |
|---|---|---|
| Deriving summary measures | CCI (Sundararajan et al., 2004) | Charlson Comorbidity Index |
|  | ECI (Mehta et al., 2018) | Elixhauser Comorbidity Index |
|  | RSI (Verdecchia, 2003) | Risk Stratification Index |
| Mapping to higher level code hierarchy or standard terminologies | ICD (Rasmy et al., 2020) | Map ICD-9 4 or 5-digit codes to ICD-9 3-digit codes |
|  | CCS (Rasmy et al., 2020) | Map ICD-9 4 or 5-digit codes to CCS codes |
| Embedding methods | Word2Vec (Tang et al., 2018) | Element-wise sum of medical concepts' embeddings generated from Word2Vec |
|  | GloVe (Tang et al., 2018) | Element-wise sum of medical concepts' embeddings generated from GloVe |
|  | FastText (Tang et al., 2018) | Element-wise sum of medical concepts' embeddings generated from FastText |

---

[2] The 4 or 5-digit ICD-9 codes are mapped to 3-digit ICD-9 codes according to the latest ICD-9 hierarchy.
[3] We use the latest version of ICD-9 to CCS single-level mapping provided by HCUP: https://hcup-us.ahrq.gov/.

| | Phe2Vec (Freitas et al., 2020) | Weighted sum of medical concepts' pre-computing embeddings |
| --- | --- | --- |
| Supervised deep learning | Med2Vec (E. Choi et al., 2016) | Supervised two-layer neural network |

Our testbed is based on the MIMIC-III database, a publicly available clinical database containing records for more than 50,000 patients admitted to the critical care units at Beth Israel Deaconess Medical Center (Johnson et al., 2016). To evaluate the performance of our method across diverse patient populations and healthcare analytical tasks, we extract four datasets that represent different patient cohorts (Table 3). MIMIC III-all dataset represents the entire patient cohorts in the MIMIC III database. MIMIC III-top15 is extracted using the top 15 most common diagnosis codes due to the long-tailed distribution of diseases in the MIMIC III database (i.e., many rare diseases are represented by only a small number of patients, while a few common diseases are represented by a large number of patients). This is typical in healthcare data and has implications for analysis and modeling. MIMIC III-top15 is used in the experiments of identifying and visualizing patient cohorts. The remaining two datasets are MIMIC III-202 and MIMIC III-572, which are rare disease cohorts for transfer learning task evaluation.

Table 3: Datasets

| Datasets | | MIMIC III-202 | MIMIC III-572 | MIMIC III-top15* | MIMIC III-all |
| --- | --- | --- | --- | --- | --- |
| Number of patient records | | 104 | 244 | 15,602 | 58,929 |
| Age | Range | 18 - 89 | 18 - 89 | 18 - 89 | 18 - 89 |
| | Mean | 71 | 59 | 68 | 55 |
| Gender | Male | 64.4% | 68.4% | 60.1% | 55.9% |
| | Female | 35.6% | 31.6% | 39.9% | 44.1% |
| Ethnicity | White | 84.6% | 70.1% | 70.4% | 69.5% |
| | African American | 3.8% | 6.1% | 7.1% | 9.2% |
| | Hispanic | 0.0% | 5.7% | 2.0% | 2.9% |
| | Asian | 1.0% | 0.0% | 1.5% | 2.6% |
| | Other / Unknown | 10.6% | 18.1% | 19.0% | 15.8% |
| Patient outcomes | Mortality | 26.0% | 40.2% | 13.2% | 9.9% |

Note: * ICD-9 codes: 41401, 0389, 41071, 4241, 51881, 431, 486, 5070, 4280, 4240, 430, 5849, 41011, 41041, 5789.

### 4.1 Evaluating medical concept-level and patient-level representations

In this section, using the MIMIC III-top 15 dataset, we evaluate the quality of the generated representations. First, we assess the alignment of *medical concept-level representations* with existing medical domain knowledge. Second, we analyze the *patient-level representations* to assess their capability of differentiating heterogeneous patient cohorts.

**4.1.1 Medical concept-level representation evaluation**

We first evaluate the quality of the medical *concept-level representations* in terms of their alignment with existing medical domain knowledge through a medical code referencing task. To serve as our gold standard, we utilize Clinical Classification Software (CCS) codes[4], which are constructed by medical experts and used to cluster medical concepts into clinically meaningful categories. The evaluation results are presented in Figure 4.

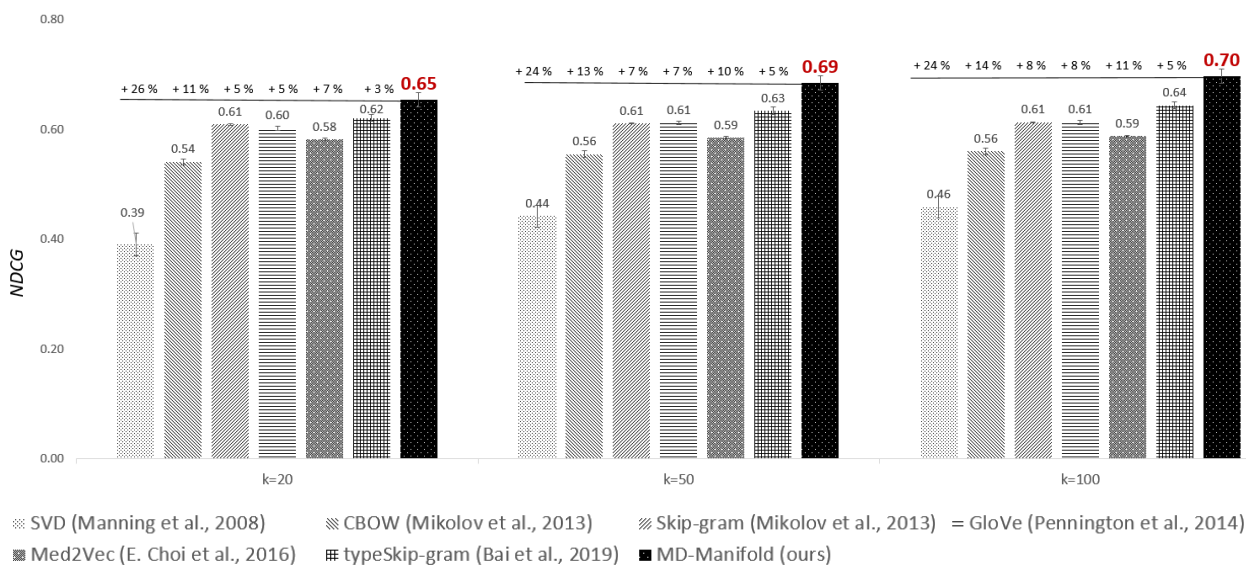

Figure 4: Comparison of medical concept representations in code referencing task

Formally, in the vector space formed by the medical concept-level representations, which includes $n$ ICD-9 codes and $m$ CCS codes $\{icd_1, ..., icd_n, ccs_1, ..., ccs_m\}$, we evaluate the quality of the generated representations by finding the closest top $k$ ICD-9 codes to each CCS code and comparing them with medical domain knowledge: $\{ccs_1: \{icd_w, ..., icd_v\}_{x_1}, ..., ccs_i: \{icd_j, ..., icd_k\}_{x_i}, ..., ccs_m: \{icd_o, ..., icd_q\}_{x_m}\}$, where $x_i$ represents the number of ICD-9 codes in the concept group $ccs_i$. For each CCS code, $ccs_i$, we find its closest top $k$ ICD-9 codes based on similarity. ICD-9 codes ranked at position $q$ is assigned label $r_q = 1$ if it is in the concept group $ccs_i: \{icd_j, ..., icd_k\}$ and $r_q = 0$ otherwise. The $NDCG@k = \frac{DCG@k}{IDCG@k}$, where $DCG@k = \sum_{p=1}^{k} \frac{r_p}{log(p+1)}$ and $IDCG@k = \sum_{p=1}^{x_i} \frac{1}{log(p+1)}$, is used to evaluate the quality of medical concept-level representations. We report the average $NDCG@k$ of $m$ CCS codes. The higher the score, the better the alignment of the generated medical concept level representations to the medical domain knowledge in ICD-9 to CCS codes mapping.

---

[4] CCS FOR ICD-9-CM, Appendix A: Single-Level Diagnoses: https://hcup-us.ahrq.gov/toolssoftware/ccs/ccs.jsp.

MD-Manifold improves the NDCG@100 score by 6% compared to the best-performing code referencing method in literature (Bai et al., 2019), providing significant value to downstream medical-concept level tasks. These representations are crucial in healthcare-related applications. For instance, MD-Manifold can find the most similar medical concepts in different ontologies (as shown in the above experiment), enabling more accurate code referencing. In healthcare-exploratory research, these representations can help researchers gain insights into medical concepts and their relationships with other concepts. By leveraging the rich information captured in the representations, researchers can perform exploratory data analysis, identify patterns, and generate hypotheses. In medical concept information retrieval, the representations can be utilized to improve the accuracy and relevance of search results. This is particularly important in healthcare, where the volume and complexity of medical data can make it challenging to find relevant information. Search algorithms can enhance the efficiency of the retrieval process by utilizing concept representations with good qualities. Overall, effective medical concept-level representations are essential in healthcare, as they enable a deeper understanding of medical concepts and support informed decision-making.

**4.1.2 Patient-level representation evaluation**

Next, we evaluate the effectiveness of patient-level representations in differentiating heterogeneous patient groups. We first generate patient-level representations and then cluster the patient representations using k-nearest neighbors algorithm. To measure the quality of the resulting clusters, we use two metrics: the ratio of inter-cluster and intra-cluster distances, $R = \frac{D_{inter}}{D_{intra}} \in [0, \infty)$, where $D_{intra}$ is average intra-cluster distance and $D_{inter}$ is the average inter-cluster distance, and the Silhouette coefficient

$S = \frac{1}{N}\sum_i \frac{b(i)-a(i)}{max(a(i),b(i))} \in [-1, 1]$, where $a(i) = \frac{1}{|C_I|-1}\sum_{j \in C_I, i \neq j} d(i,j)$ denotes the mean distance between

$i$ and all other data points in the same cluster, and $b(i) = min_{J \neq I} \frac{1}{|C_J|}\sum_{j \in C_J} d(i,j)$ denotes the minimum

mean distance of $i$ to all points in any other cluster. The results are presented in Figure 5. Higher $R$ and $S$ values suggest that the patients within a patient group are tightly clustered and well-separated from other patient groups. This indicates that the patient groups are homogeneous and distinguishable from each other. Conversely, lower $R$ and $S$ values suggest that the patient records within a patient group are widely dispersed, or there is significant overlap with other patient groups. This indicates that the patient groups are heterogeneous and difficult to distinguish from each other, as shown in Figure 5 (b).

Figure 5: Comparison of patient representations in distinguishing patient cohorts
(a) Performance improvements compared to other patient-level representations

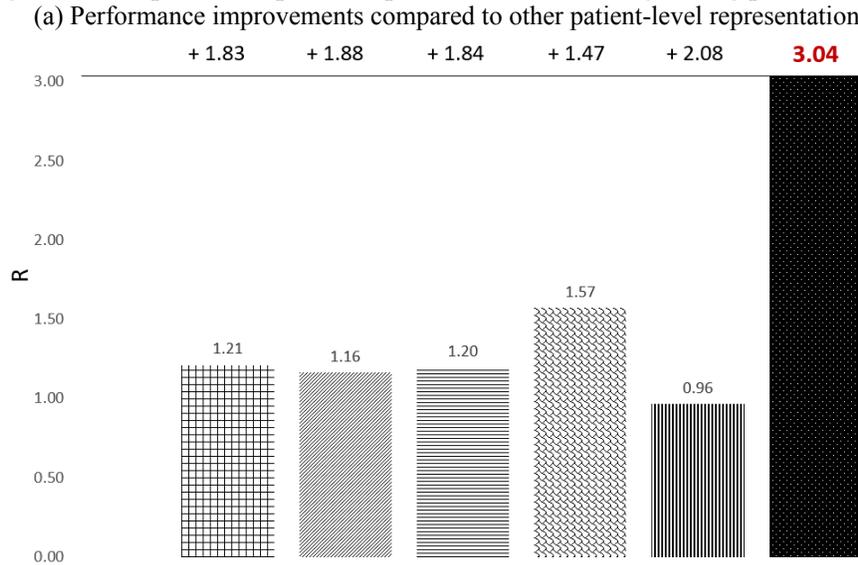

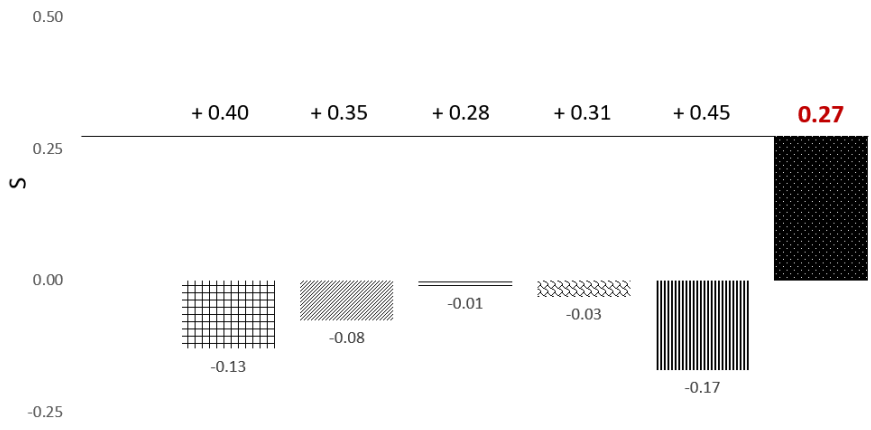

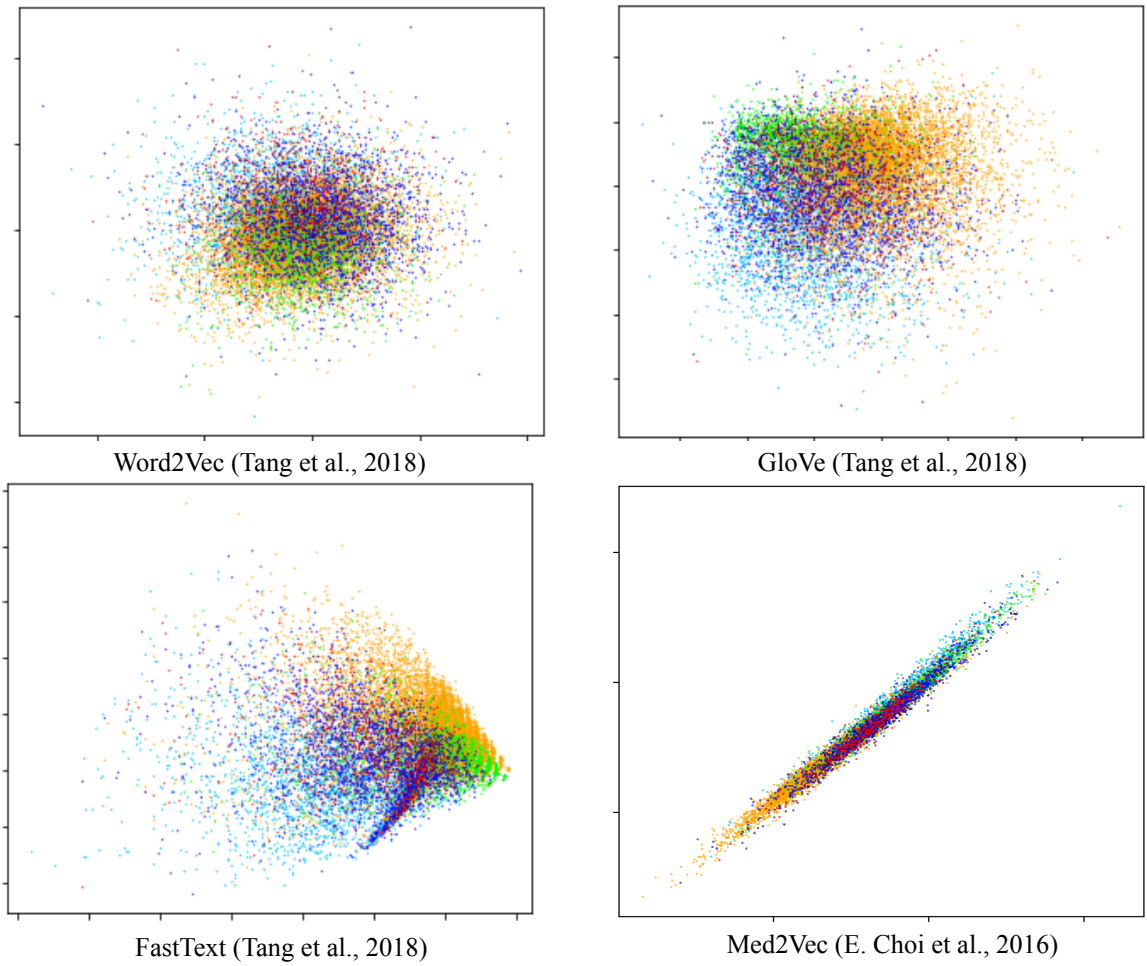

(b) Visualization of patient representations in distinguishing patient cohorts

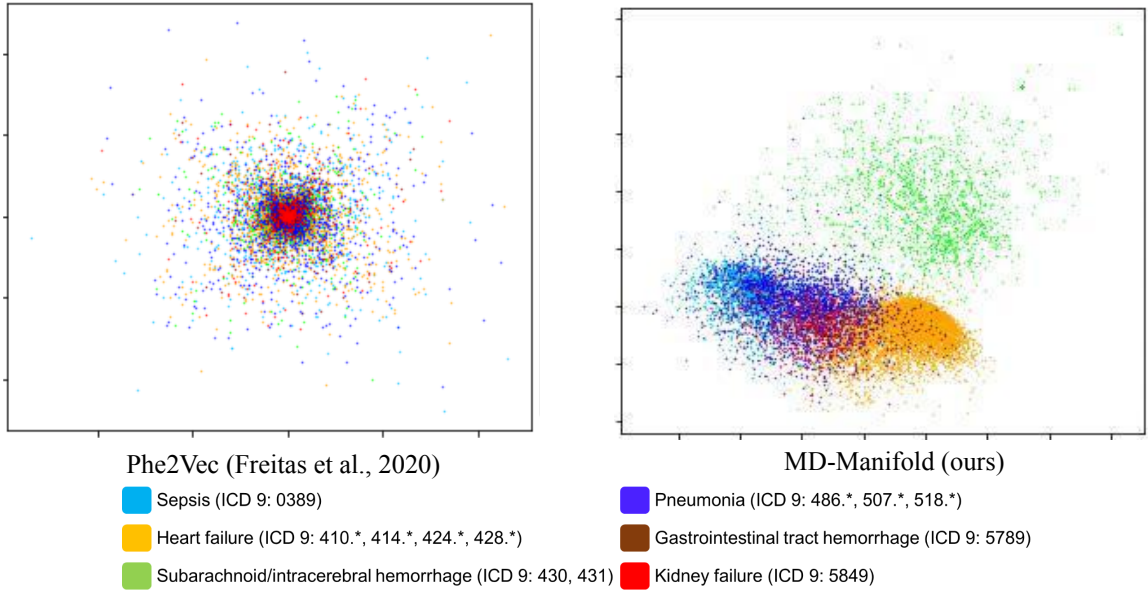

| Phe2Vec (Freitas et al., 2020) | MD-Manifold (ours) |

- 🟦 Sepsis (ICD 9: 0389)
- 🟪 Pneumonia (ICD 9: 486.*, 507.*, 518.*)
- 🟧 Heart failure (ICD 9: 410.*, 414.*, 424.*, 428.*)
- 🟫 Gastrointestinal tract hemorrhage (ICD 9: 5789)
- 🟩 Subarachnoid/intracerebral hemorrhage (ICD 9: 430, 431)
- 🟥 Kidney failure (ICD 9: 5849)

Our method has demonstrated significant improvement over the current state-of-the-art patient-level representations in distinguishing heterogeneous patient groups. This capability is crucial for healthcare analytical tasks. First, it enhances the accuracy and personalization of health condition analysis for specific patient populations, enabling more precise patient cohort selection and patient summarization using medical concepts. Second, patient groups may exhibit diverse clinical characteristics, disease progression, and treatment responses that can affect the effectiveness of healthcare interventions. By using patient representations that can accurately differentiate between patient groups, healthcare analytical algorithms can identify patterns and relationships that may not be apparent in the general patient population. Therefore, patient-level representations that can distinguish heterogeneous patient groups can significantly improve the performance of downstream healthcare analytical tasks.

**4.2 Downstream healthcare analytical applications using patient-level representations**

Learning compact and effective representations is crucial not only for obtaining descriptive insights into medical concepts and patient-level tasks but also for improving predictive foresight for various downstream healthcare-analytical tasks. In this section, we demonstrate the effectiveness of our generated patient representations, which can significantly enhance the performance of patient outcome predictions, transfer learning for rare disease cohorts, and multimodal medical data fusion. We use ICU patients' in-hospital mortality prediction as the prediction task to evaluate the effectiveness of our proposed method. ICU patients' mortality prediction is of paramount importance for assessing the severity of disease, adjudicating new treatments, comparing patients' cohorts treated across different hospitals, allocating resources and determining levels of care, and discussing expected outcomes with the hospitalized patients (Pirracchio et al., 2015). Using ICU patients' mortality prediction as a research case, we aim to show that the patient representations generated by our method can effectively incorporate medical domain knowledge and prior data information, and therefore enhance the performance of downstream healthcare analytical tasks.

**4.2.1 Enhancing ICU patient mortality prediction through medical code representation**

Accurate patient representations can enhance machine learning-based patient outcome predictions by reducing noise and irrelevant information, thereby enabling more precise pattern recognition and predictions. We examine the performance of prediction models using our patient representations compared to state-of-the-art patient representations on the MIMIC III-all dataset.

Figure 6 (a) presents a comparison between MD-Manifold and traditional feature engineering methods for patient representations. Results show that MD-Manifold outperforms the best-performing method that employs summary measures to represent patients (Mehta et al., 2018; Sundararajan et al., 2004; Verdecchia, 2003), yielding a 9% increase in AUC score. This performance gap can be attributed to the oversimplification of patient information by summary measures, which collapse various patient characteristics and conditions into a single value, resulting in a loss of important details and nuances that may be relevant for accurate healthcare outcome prediction. Furthermore, summary measures may not adequately capture the heterogeneity of patient cohorts, leading to inaccurate predictions. In comparison to methods that map medical concepts to higher-level code hierarchies or standard terminologies for representation (Rasmy et al., 2020), MD-Manifold exhibits a better performance, improving the AUC score by 12% and 5%, respectively. This performance gap may arise due to the loss of specificity in the representation of the patient's medical condition by these methods, as the mapping process can discard unique details and nuances present in the original medical concepts. Additionally, these methods may fail to capture essential dynamic prior information in patient records, such as comorbidities, which are crucial for healthcare outcome prediction.

Figure 6 (b) presents a comparison of MD-Manifold with embedding-based methods for patient representation across different dimensions. The results indicate that MD-Manifold outperforms the best-performing embedding-based method (Tang et al., 2018) by 1.0% (dimension=64), 1.8% (dimension=128), and 1.8% (dimension=256) across various dimensions of patient representation in terms of AUC scores. While embedding-based methods are powerful in capturing semantic and syntactic relationships between words, MD-Manifold still exhibits improved performance in healthcare outcome prediction tasks. The better performance of MD-Manifold can be attributed to its ability to capture the underlying medical knowledge and relationships of the medical concepts used for patient representation. In contrast, embeddings may not fully capture this knowledge, leading to a loss of important medical domain information that could be critical for healthcare outcome prediction.

While we are only presenting the results of mortality prediction, our patient representation can also be applied to predict other healthcare outcomes, such as readmission rates, length of stay, effectiveness of care, and more. The purpose of this experiment is not to propose a comprehensive solution for ICU patient outcome prediction. We use a simple classifier structure (Appendix B) and only include medical concepts from structured data as input, while there are more complex model structures and many other types of medical data available. Our aim is to demonstrate the significance of fully representing and mining medical concepts. Overall, the results suggest that MD-Manifold offers performance improvements over feature engineering and embedding methods for patient representation using medical concepts, making it a promising tool for downstream healthcare outcome predictions.

Figure 6: Comparison of patient representations in ICU mortality prediction task
(a) Comparison with feature engineering methods

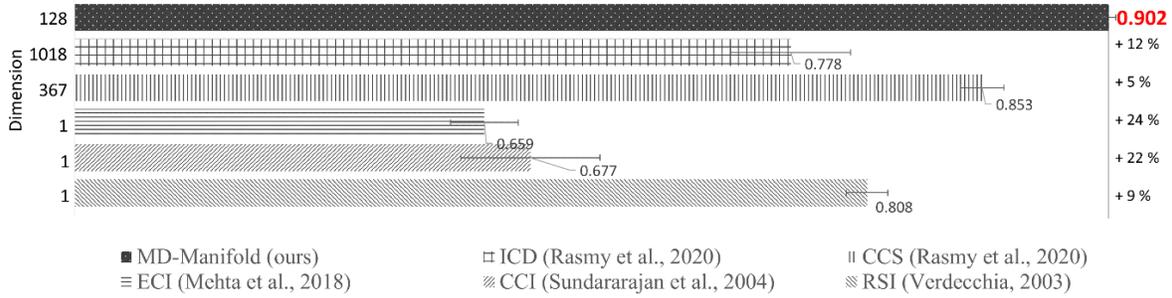

(b) Comparison with embedding methods

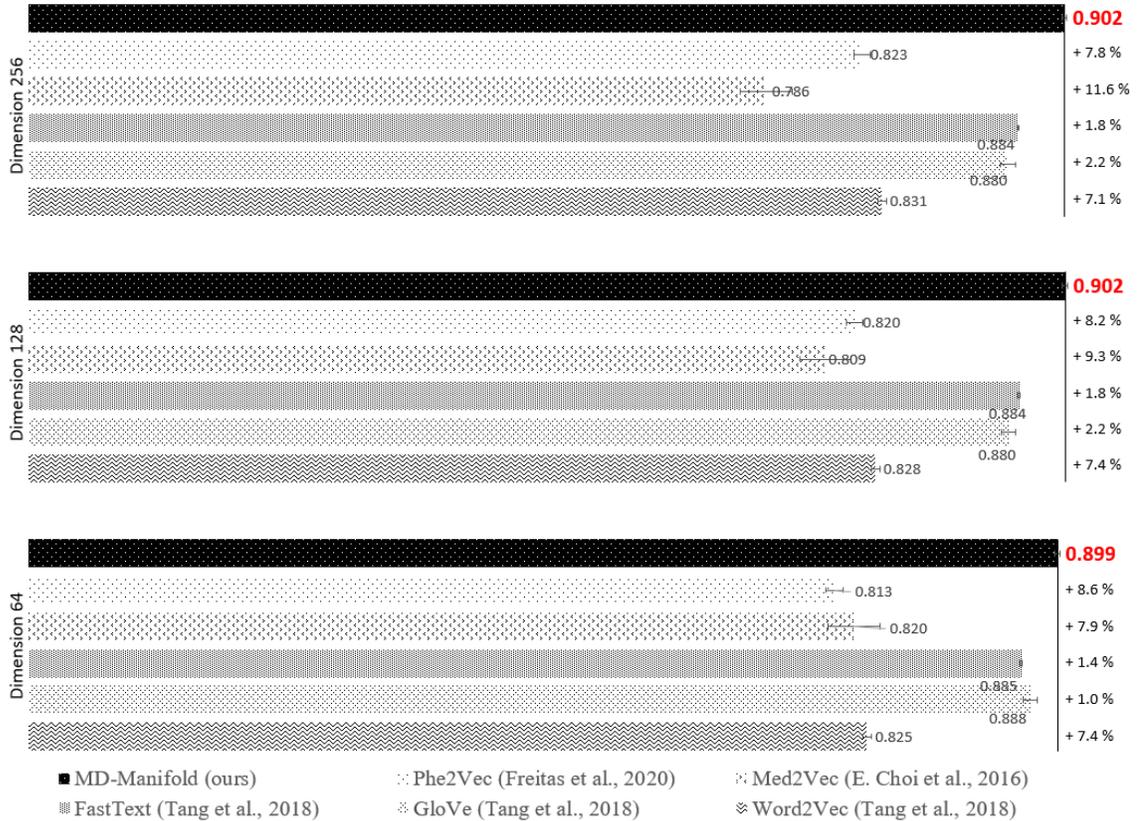

### 4.2.2 Enhancing transfer learning for ICU mortality prediction in rare disease cohorts through medical code representation

Transfer learning involves leveraging knowledge learned from a source domain to improve performance in a target domain. If the source domain is similar to the target domain, machine learning models can leverage the learned knowledge from the source domain to perform better on the target domain. This is particularly important when the target domain has limited data or a different distribution from the source domain. In healthcare, transfer learning can be particularly useful for predicting outcomes of rare diseases, where patient records are often scarce and challenging to obtain. By using good patient representations and similarity calculation, we can identify similarities between rare disease patients and other patients. By adding these patients to the rare disease cohorts, we can increase the number of training data records, which may lead to more accurate predictions of patient outcomes and better overall patient care.

Figure 7: Comparison of patient representations for ICU mortality prediction in rare disease cohorts

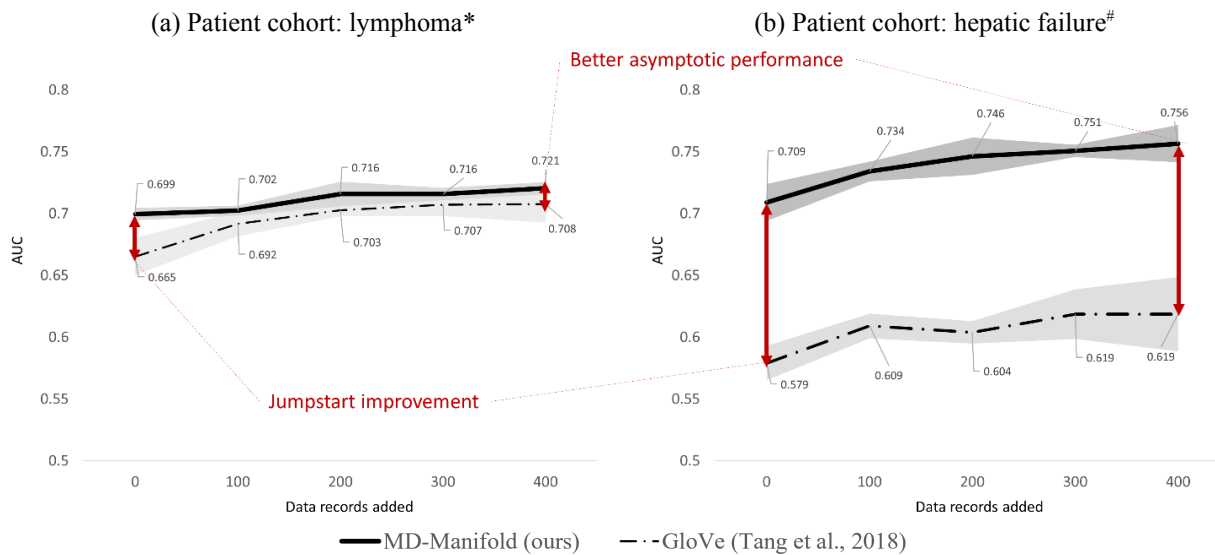

Note: * ICD-9 code: 202, number of patients: 104; # ICD-9 code: 572, number of patients: 244

We investigate the use of transfer learning for rare disease outcome prediction on the MIMIC III-202 and MIMIC III-572 datasets, which have only 104 and 244 patient records, respectively. This poses a challenge for machine learning models, particularly deep learning models, as their performance tends to suffer with such limited data. Using patient-level representations, we propose a straightforward way to conduct transfer learning for rare disease outcome prediction. Formally, given a rare disease patient cohort as the target domain, $D_T = \{Y_T, X_T\}$, our goal is to improve the prediction performance of $Y_T$. Using patient representations, we generate a source domain, $D_S = \{Y_S, X_S\}$, by selecting the top $n$ ($n \in \{0, 100, 200, 300, 400\}$) most similar patients to $X_T$. $D_T$ is split into $D_{T\_train}$, $D_{T\_validation}$, and $D_{T\_test}$. The machine learning models are trained on $\{Y_S + Y_{T\_train}, X_S + X_{T\_train}\}$ and validated and tested on $\{Y_{T\_validation}, X_{T\_validation}\}$ and $\{Y_{T\_test}, X_{T\_test}\}$, respectively.

Experimental results show that, compare to the best-performing patient representations in section 4.2.1 (i.e., GloVe (Tang et al., 2018)), our patient representations significantly improve transfer learning prediction performance and demonstrate jumpstart improvement and better asymptotic performance on both rare disease cohorts when utilized to generate the source domain (Figure 7). The performance gain of the jumpstart improvement can be attributed to our patient representations capturing key features unique to the rare disease, which can be difficult to identify through other methods. Meanwhile, our approach may have contributed to better asymptotic performance because we incorporate medical domain knowledge and prior data information to create patient representations. By doing so, we are able to identify the most similar patients and generate the most meaningful $D_S$, resulting in improved health outcome predictions for rare disease cohorts. This approach has the potential to accelerate progress towards personalized medicine and better outcomes for patients with rare diseases. The experimental results also demonstrate the potential of our method for other IS problem domains where labeled data is scarce and challenging to obtain, thereby helping researchers effectively conduct transfer learning on such problem domains.

The purpose of this experiment is not to propose a new transfer learning solution for rare disease patient cohorts as the field of transfer learning is constantly evolving, with new approaches emerging all the time. However, in general, transfer learning heavily relies on the choice of source domain. Our aim is to demonstrate that better patient representation can facilitate the selection and generation of the source domain, and thus potentially improve the performance of transfer learning tasks.

### 4.3.3 Enhancing multimodal data fusion through medical code representation

Medical data is one of the most complex types of data due to its mixture of structured data (such as medical records and demographic data) and unstructured data (such as medical images, clinical notes, and vital sign series). Multimodal healthcare data fusion involves integrating data from different sources to provide a comprehensive understanding of a patient's health status. While existing solutions from computer or data science may suffice for other data modalities' representation and fusion (Alsentzer et al., 2019; Shen et al., 2017; Song et al., 2018), medical concepts have unique characteristics, such as rich domain knowledge and prior data information, that require in-depth study.

We evaluate the potential of our medical concept representations to improve an existing framework[5] (Nguyen et al., 2019) that utilizes patients' demographic, clinical notes, lab results, and vital sign data for ICU outcome prediction. Specifically, we fuse the backbone model's input with patient representations (Fukui et al., 2016), which we obtain solely from medical codes. We then compare the prediction results of using our representation and the best performing patient representations from section 4.2.1 (i.e., GloVe (Tang et al., 2018)) on the MIMIC III-all dataset.

Figure 8: Comparison of patient representations for data fusion task

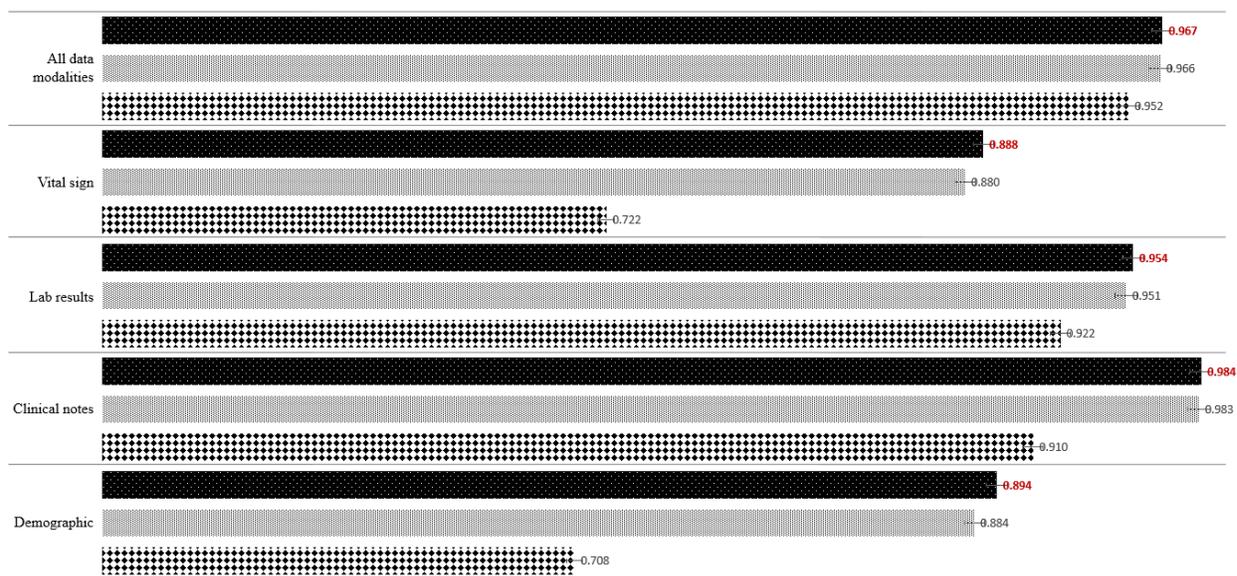

The results demonstrate that our model can effectively utilize the potential of medical codes for ICU outcome prediction across all data modalities, outperforming the backbone model. Specifically, we observed AUC improvements of 18.6% on demographic data, 7.4% on clinical notes, 3.2% on lab results, 16.6% on vital signs, and 1.5% across all data modalities. Furthermore, when compared to state-of-the-art patient representations (Tang et al., 2018), our representation still shows better performance. While the performance gain is not significant, our method still has an advantage over Tang et al.'s (2018) approach.

---

[5] We chose this model as the backbone because of two reasons: (1) its data and code are publicly available on the internet, making it easy to replicate; (2) it uses multiple different modalities of medical data, excluding medical codes, making it an ideal backbone model to test the added value of medical concept representations.

Their method is based on deep learning techniques, where the model was discriminatively trained to learn a conditional distribution of outputs given inputs. For instance, the model predicted a medical concept given other medical concepts in the same patient record or predicted other medical concepts in the same patient record given a medical concept. On the other hand, manifold learning algorithms act as generative models, which aim to capture the actual distribution of the data. We anticipate that generative models would perform better with less training data, whereas discriminative models would catch up with sufficient training data (Ng & Jordan, 2001). As a significant amount of training data is not always available for healthcare prediction tasks (E. Choi et al., 2018), using manifold learning algorithms provides an advantage to the proposed method (as demonstrated by the results of transfer learning experiments.) It is important to note that the level of information richness and the degree of overlap with medical concept representation vary across different data modalities, resulting in varying levels of performance improvement. Moreover, it is also necessary to recognize that not all data modalities are always available in various medical data. Nevertheless, medical concepts are widely present in medical data and fully exploiting the potential of medical concepts can improve the performance of downstream healthcare outcome prediction tasks.

The objective of this experiment is not to present a comprehensive solution for multi-data fusion, but rather to showcase, through a simple example of multi-data fusion, that thoughtful medical concept representations can enhance the predictive performance of downstream tasks. Multimodal data fusion is a distinct field with rapid development and numerous publications, but, in general, all fusion techniques depend on effective data representations. We believe that medical concept representation is currently a bottleneck in medical data representation. Our approach may pave the way for better medical data fusion techniques, as these data representations can be utilized by the latest advances in multimodal data fusion research.

**5. Discussion and future research**

The rise of HealthIT has resulted in the availability of a vast amount of medical data. Among various types of medical data, medical concepts possess unique characteristics, including high-dimensionality and significant domain-specific knowledge or prior data information. This presents researchers with both an opportunity and a challenge - to represent medical concepts and patients effectively for healthcare analytical applications. In accordance with the design science research paradigm (Gregor & Hevner, 2013; Hevner et al., 2004), this study proposes a novel approach to data augmentation that leads to the creation of a new representation of medical concepts. Building on these new concept-level representations, we introduce a new medical distance metric, which is then used to generate new patient-patient networks that preserve critical medical knowledge embedded within the hierarchical structure of the medical concepts and patient data. Finally, we leverage manifold learning algorithms to develop new patient representations. The incorporation of the new medical distance metric and patient-patient network helps capture rich domain knowledge and prior information in medical concepts and patient records, which is then used to generate improved patient representations. Empirical evaluations using real-world datasets demonstrate that our proposed framework, MD-Manifold, generates concept-level representations that accurately represent medical knowledge. Moreover, building upon the concept-level representations, the generated patient-level representation can accurately distinguish between heterogeneous patient cohorts. MD-Manifold also outperforms other state-of-the-art techniques in various downstream healthcare analytical tasks.

**5.1 Research contributions and their implications to IS and healthcare analytical applications**

From a design science perspective, we present two contributions. First, we propose a novel framework for generating representations of medical concepts and patients that leverages existing medical domain knowledge and essential prior data information. Second, we introduce two innovative IT artifacts as part of our framework: (1) a novel data augmentation approach, distance metric, and patient-patient network that integrates critical domain knowledge and prior data information, and (2) a comprehensive framework that integrates both domain knowledge and prior data information for representation learning.

Our work also has significant implications to IS research. (1) First, our framework addresses the challenges associated with medical concepts and patient representations in healthcare analytics. Medical data is one of the most complex types of data, comprising both structured and unstructured data and containing rich domain knowledge and prior information. One area that previous research has not paid enough attention to is medical concept representation. Our research demonstrates that, in addition to designing different model structures, emphasis can also be placed on insightful data representation to further enhance the performance of healthcare explorative and predictive tasks. Our approach to representation learning proposes that, rather than adhering to the common practice of relying on medical knowledge or following computer science methodologies that use massive data to represent medical concepts, IS researchers can introduce or design new artifacts that incorporate crucial domain knowledge and prior information required for downstream tasks. Our framework represents a contribution to the IS knowledge base, as it addresses a significant application domain with existing limitations (Gregor & Hevner, 2013). (2) Second, our proposed framework highlights the significance of knowledge-driven machine learning, which involves integrating medical domain knowledge into representation learning. As machine learning models continue to grow larger and require extensive amounts of training data, it can be challenging for researchers with limited computational resources to keep up. Knowledge-driven machine learning presents a promising area of study within the field of IS, offering numerous opportunities and a viable alternative to deep and large models. By incorporating domain knowledge to develop more efficient artifacts, knowledge-driven machine learning can be particularly valuable in complex learning tasks with limited training resources. Our modeling framework and design principles offer a nascent design theory that can inspire other scholars to explore the potential of knowledge-driven machine learning (Abbasi et al., 2016; Chau et al., 2020; Yang et al., 2022; Zimbra et al., 2018). (3) Third, our study, situated within computational design science research, contributes to creating middle-ground frameworks, as proposed by Yang et al. (2022), by generating data representations that can be effectively utilized in a wide range of downstream explanatory or predictive tasks. As unstructured, complex, and high-dimensional data become increasingly prevalent, extracting valid information from such data for IS research becomes challenging. Our proposed method of generating concept representations is both knowledge-driven and data-driven, making it useful not only in medical concept representation but also in other representation learning tasks. For example, in Finance, Industrial Classification Codes (ICC) are crucial for many business practices, such as identifying potential customers and competitors. There are clear hierarchical relationships between ICC codes as well as dynamic data information (Hoberg & Phillips, 2018; Zhao et al., 2022). Our approach may balance domain knowledge with prior data information to better represent ICC codes for downstream analysis. Likewise, in knowledge management studies such as legal knowledge management and sharing, the hierarchical structure of legal terms is often based on domain knowledge, but these terms also have co-occurrences in legal cases (Breuker et al., 2002). Our proposed method can potentially offer a more effective representation of legal terms for downstream analyses by considering both the hierarchical structure and co-occurrence information in the data. To summarize, our approach has the potential to facilitate the development of impactful design artifacts in other IS problem domains.

Practically, the goal of healthcare analytics is to improve medical outcomes by using medical data to gain insights into patients and diseases. One of the key challenges in this field is to create effective representations of medical concepts and patients that can be utilized across various analytical tasks and platforms. It is vital because representations that are specific to a particular task may be limited in their scope and applicability, making it challenging to develop new analytical applications or adapt existing ones to new contexts. Our method provides a way to create more generalized representations that incorporate a significant amount of medical domain knowledge and prior medical data information, which can be used across a variety of healthcare analytical applications. By developing representations that are less task-specific, it may be possible to increase the ease of applicability of healthcare analytics and facilitate the development of new applications. This, in turn, can contribute to better healthcare outcomes by enabling researchers and practitioners to gain deeper insights into medical concepts and patient behavior, leading to more effective treatments and interventions.

**5.2 Limitations and future work**

While this work holds promise, there are several areas that could be improved upon. First, this study aims to represent a patient's health status during a medical event using medical concepts from a single patient record. However, our research can be expanded to include all medical concepts from multiple patient records, representing the patient's health status over an extended period. These patient representations can also be utilized to demonstrate changes in a patient's health over time. Second, this work primarily focuses on medical codes within structured medical data. However, medical concepts are also prevalent in unstructured data, such as clinical notes. By utilizing NLP preprocessing techniques such as named entity recognition, we can extract these medical concepts. Then, by incorporating medical domain knowledge and prior data information, our work has the potential to enhance the representation of unstructured free-text medical data. Third, to showcase the practicality of MD-Manifold, we use medical record data (i.e., MIMIC III) as the testbed. Nevertheless, the proposed method can be extended to any patient record data containing complex medical concepts, such as claims data, disease registries, and pharmaceutical data.

## 6. Conclusion

The selection of data representation is crucial for the performance of machine learning and their efficacy in tackling real-world problems. This study underscores the importance of data representation, as effectively representing complex, high-dimensional concepts remains a major challenge in various problem domains, and good data representations should be task-agnostic and capture general prior knowledge and disentangle different explanatory factors of the data. Our proposed framework takes a step towards integrating essential domain knowledge and prior data information to generate data representations that can be efficiently used in various downstream predictive or explanatory tasks. Our method and design principles can be generalized to other IS problem domains, resulting in more accurate and reliable descriptive and predictive analyses, ultimately enhancing the impact of IS research. We encourage future research to explore new approaches to representation learning and continue to make advancements in this area.

**Appendix A. Pilot study: comparison of model components**

Our proposed framework, MD-Manifold, consists of several components, including the manifold learning algorithm, the concept-level and record-level distance functions (including patient-patient networks with different distance functions). For each of these components, we have multiple choices, and as documented in the literature, each of these choices has its advantages and disadvantages. Therefore, we conduct multiple pilot experiments to determine the model details for the proposed framework. We extract four smaller datasets by using the most common diagnosis codes in the MIMIC III database for the pilot experiments (Table A.1). In all the pilot experiments, we use ICU patients' in-hospital mortality prediction as the prediction task.

Table A.1. Pilot study datasets

| Datasets | | MIMIC III-428 | MIMIC III-41401 | MIMIC III-389 | MIMIC III-41071 |
|---|---|---|---|---|---|
| Number of patient records | | 1488 | 3498 | 2069 | 1751 |
| Age | Range | 18 - 89 | 18 - 89 | 18 - 89 | 18 - 89 |
| | Mean | 72 | 67 | 69 | 71 |
| Gender | Male | 54.3% | 75.9% | 52.5% | 62.4% |
| | Female | 45.7% | 24.1% | 47.5% | 37.6% |
| Ethnicity | White | 70.2% | 68.2% | 73.2% | 69.6% |
| | African American | 14.0% | 2.8% | 10.1% | 3.5% |
| | Hispanic | 2.4% | 1.7% | 2.1% | 1.4% |
| | Asian | 0.8% | 1.4% | 2.0% | 0.7% |
| | Other / Unknown | 12.7% | 25.8% | 12.6% | 24.8% |
| Patient outcomes | Mortality | 12.2% | 8.9% | 31.6% | 8.0% |

**Appendix A.1. Manifold learning algorithms**

This section compares two different types of manifold learning algorithms: Laplacian Eigenmap and Isomap. (1) Laplacian Eigenmap first computes nearest neighbors for each data point and creates the weighted nearest neighbor network (i.e., node: each data point, edge: nearest neighbor relationship, edge weight: proportional to the reverse distance between nearest neighbors). The larger the edge weight, the more similar the nodes, and the closer they are in the manifold space. Laplacian Eigenmap computes a low-dimensional representation of the data point and optimally preserves the local neighborhood information. (2) Isomap also calculates the nearest neighbors for each data point and develops the nearest neighbor network (i.e., node: each data point, edge: nearest neighbor relationship, edge length: distance between nearest neighbors). It then computes the shortest path distances between all pairs of points in the network. Isomap preserves the global structure of the original manifold space and finds the optimum low-dimensional representation for data points by retaining the geodesic distance between each pair of nodes on the constructed nearest neighbor network.

In the pilot experiment, generally, the Isomap has a better performance (see Table A.1.1). The highest AUC scores of Isomap at different dimensions are usually higher than that of Laplacian Eigenmap. The possible reason is that Isomap is more robust to noise than Laplacian Eigenmap; similar findings are also reported by Mysling et al. (2011) and Talwalkar et al. (2013). We only report Isomap's performance in the evaluation section for brevity.

Table A.1.1: ICU in-hospital mortality prediction using different manifold learning algorithms (AUC)

| Dataset | Classifier | Dimension | | | | | |
|---|---|---|---|---|---|---|---|
| | | 16 | 32 | 64 | 128 | 256 | 512 |
| MIMIC - 0389 | Isomap | 0.743 | 0.789 | **0.816** | 0.814 | 0.788 | 0.732 |
| | Eigenmap | 0.739 | 0.773 | 0.798 | 0.802 | 0.778 | 0.728 |
| MIMIC - 428 | Isomap | 0.754 | 0.781 | 0.782 | **0.783** | 0.737 | 0.657 |
| | Eigenmap | 0.752 | 0.763 | 0.765 | 0.778 | 0.731 | 0.668 |
| MIMIC - 41071 | Isomap | **0.880** | 0.868 | 0.847 | 0.836 | 0.781 | 0.752 |
| | Eigenmap | 0.877 | 0.872 | 0.839 | 0.827 | 0.783 | 0.769 |
| MIMIC - 41401 | Isomap | 0.853 | **0.921** | 0.910 | 0.846 | 0.809 | 0.797 |
| | Eigenmap | 0.803 | 0.859 | 0.865 | 0.835 | 0.855 | 0.802 |

Note: The best performance on each dataset is bold.

**Appendix A.2 Distance metrics and patient-patient networks**

In this section, we evaluate the performance of different patient-patient networks, $G$, generated using different combinations of medical concept-distance metrics (i.e., benchmark metric[6] $CD_{WP}$ (Wu & Palmer, 1994) and our proposed metrics $CD_{new}$, including $CD_{new-Cosine}$, $CD_{new-Manhattan}$, $CD_{new-Euclidean}$, and $CD_{new-eHDN}$) and medical record-distance metrics (i.e., four widely used metrics $SD_1$, $SD_2$, $SD_3$, and $SD_4$). n_neighbors is a hyperparameter of $G$, determined through grid search.

---

[6] We adopt Wu and Palmer (1994)'s metric as the baseline on account of its simple design and powerful performance (Jia et al., 2019).

The experimental results are reported in Figure A.2.1 and they reveal several interesting findings. (1) The red dotted lines in Figure A.2.1 represent the best performance on each dataset. On all four datasets, the performance of $CD_{WP}$ (gray lines) never achieve the top AUC scores in both prediction tasks. Therefore, our metrics $CD_{new}$ are more effective at measuring the distances between medical concepts, resulting in higher AUC scores in healthcare prediction tasks. (2) We propose a new medical-concept distance metric $CD_{new}$ with four distance formulas: $Cosine$, $Manhattan$, $Euclidean$, and $eHDN$. In our evaluation, $CD_{new-Cosine}$ and $CD_{new-eHDN}$ outperform the $CD_{new-Euclidean}$ and $CD_{new-Manhattan}$. $CD_{new-Euclidean}$ and $CD_{new-Manhattan}$ never achieve the highest AUC scores on all four datasets. Especially, when paired with $SD_2$, $CD_{new-Cosine}$ is the best $CD$ metric (AUC = 0.783) for the in-hospital mortality prediction on the MIMIC III - 428 dataset (Figure A.2.1). This is an interesting finding because it indicates that it is important to normalize the co-occurrences of medical concepts for medical-concept distance calculation when considering disease co-occurrences as medical domain knowledge. Both $CD_{new-Cosine}$ and $CD_{new-eHDN}$ include normalization terms in the distance formulas (i.e., $\sqrt{C_a \cdot C_a}\sqrt{C_b \cdot C_b}$ and $\sqrt{\Sigma C_a \Sigma C_b (N - \Sigma C_a)(N - \Sigma C_b)}$). The significance of the normalization terms is that they eliminate the impact of very popular diseases across all patient cohorts. For example, a very popular medical concept $M_j$ co-occurs with most other medical concepts. Therefore, most of the elements in row $j$ in the co-occurrence matrix $C$ are large values. By contrast, there are two rare medical concepts $M_a$ and $M_b$. The elements in both rows $a$ and $b$ are small numbers in the co-occurrence matrix $C$. If $M_a$ and $M_b$ co-occurs frequently, we expect the medical-concept distance to reflect such a co-occurring relationship. However, $CD_{new-Euclidean}$ and $CD_{new-Manhattan}$ may fail to capture such a pattern, leading to an undesired performance in healthcare prediction tasks. This finding echoes other studies which show the significance of co-occurrence normalization (Kumar et al., 2015).

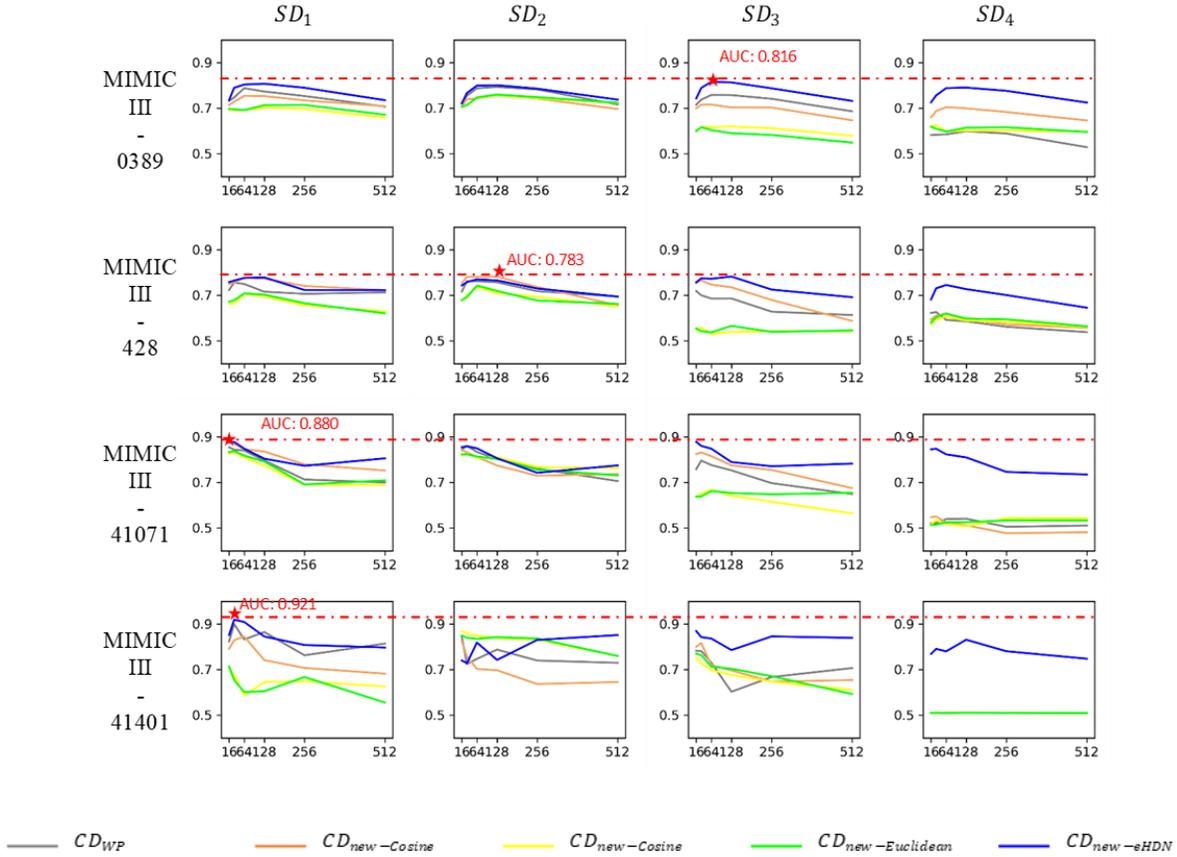

Figure A.2.1: Performance of different distance functions and patient-patient networks

Note: (1) Medical knowledge: $T_{ICD9}$. (2) The orange, gray and yellow lines are overlapped by the green line in the line charts on the MIMIC III 41401 dataset using $SD_4$.

(3) We also evaluate the performance of four distance metrics, $SD = distance(V_i, V_j)$, for measuring the distances between medical records. $SD_1$ shows better performance compared to other medical record distance metrics. The possible explanation is that $SD_1$ is designed to capture the similarities of the most similar medical-concept pairs from two medical records, which are essential features for the two healthcare prediction tasks. An interesting finding is that $SD_4$ is also developed to compare the most similar medical-concept pairs from two medical records. However, $SD_4$ performs poorly in the prediction tasks on all four datasets. As shown in Figure A.2.1, $SD_4$ never achieves the best AUC scores. This result differs from the finding of Jia et al. (2019). The difference between $SD_1$ and $SD_4$ lay in how they define the most similar medical-concept pairs (see Figure A.2.2). In $SD_1$, every medical concept can be paired with another medical concept. Such a pair forms a set of "most similar pairs" for medical-record distance calculation. However, in $SD_4$, it is possible that a medical concept is not paired with other medical concepts. Hence, $SD_4$ excludes such a medical concept from medical-record distance calculation, which compromises the accuracy of downstream tasks. The experimental results suggest that every medical concept contains important information like disease diagnosis and is important for healthcare analytical tasks.

Figure A.2.2: Example of most similar medical-concept pairs in $SD_1$ and $SD_4$

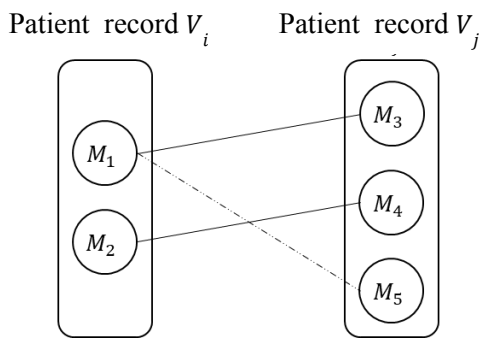

Most similar pairs in $SD_1$ (solid and dash lines): $<M_1, M_3>, <M_2, M_4>, <M_1, M_5>$.
Most similar pairs in $SD_4$ (solid lines): $<M_1, M_3>, <M_2, M_4>$.

To summarize, we develop a new medical concept-distance metric $CD_{new}$ that is both knowledge-driven and data-driven to preserve medical domain knowledge in medical concepts' properties, i.e., the hierarchical structure and co-occurrences. Extant metric $CD_{WP}$ does not consider the co-occurrences of medical concepts, hence is outperformed by our metric $CD_{new}$. Since $CD_{new}$ take medical concepts' co-occurrences into consideration, it is important to use distance formulas with normalization terms that normalize the co-occurrences of medical concepts. For brevity, we only report the results of using the patient-patient network generated from $CD_{new-Cosine}$ and $SD_1$ in the evaluation section.

**Appendix A.3 Medical domain knowledge as the prefix trees**

In this section we explore the performance of three prefix trees as medical domain knowledge, i.e., $T_{ICD9}$, $T_{CUI}$, and $T_{CCS}$. Specifically, (1) $T_{ICD9}$ represents the relationship between a medical concept and its higher-level ICD-9 disease diagnosis categories, as shown in Figure A.3.1 (a). ICD-9 diagnosis codes are composed of codes with 3, 4, or 5 digits, which are all medical concepts. Three-digit ICD-9 codes stand for the categorical information of diseases. Three-digit ICD-9 codes are further divided by the use of fourth and/or fifth digits, which provides greater details of diseases. Hence, the medical domain knowledge is contained in the ICD-9 diagnosis codes' hierarchical structure. (2) $T_{CUI}$ exhibits the relationship between the medical concepts and the corresponding Concept Unique Identifiers (CUI) from the UMLS. As shown in Figure A.3.1 (b), two medical concepts may indicate similar diagnoses, and CUI links these medical concepts in $D$ that mean exactly or nearly the same. Therefore, UMLS Metathesaurus structure, which represents the properties of diseases and their relations to other diseases, serves as the source of medical domain knowledge. (3) $T_{CCS}$ reflects the projection of medical concepts $M_j$'s onto the CCS categorization scheme. As the example in Figure A.3.1 (c) shows, a group of medical concepts at the bottom can be collapsed into a smaller number of clinically meaningful categories (CCS codes) that may be useful for presenting descriptive statistics than the individual medical concept.

Figure A.3.1: Examples of medical-concept hierarchy structures

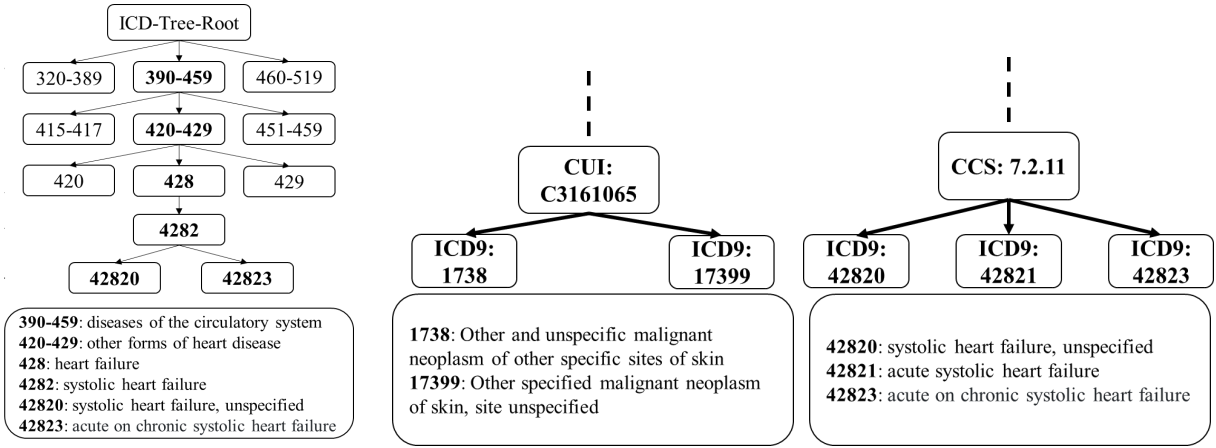

(a) An example of $T_{ICD-9}$  (b) An example of $T_{CUI}$  (c) An example of $T_{CCS}$

Our results (Table A.3.1) suggest that using proper medical domain knowledge can provide useful information for medical concept distance calculation, which is consistent with Melton et al. (2006)'s finding. As $T_{ICD9}$, $T_{CUI}$, and $T_{CCS}$ exhibit comparable performance, we only present the evaluation results from $T_{ICD9}$ in the evaluation section to keep the presentation concise.

Table A.3.1: ICU mortality prediction performance (AUC) using different medical domain knowledge

| Dataset | MIMIC III-389 | | | MIMIC III-428 | | | MIMIC III-41071 | | | MIMIC III-41401 | | |
|---|---|---|---|---|---|---|---|---|---|---|---|---|
| Dimensions | 64 | 128 | 256 | 64 | 128 | 256 | 64 | 128 | 256 | 64 | 128 | 256 |
| $T_{ICD9}$ | **0.82** | 0.81 | 0.79 | 0.78 | 0.78 | 0.74 | 0.85 | 0.84 | 0.78 | **0.91** | 0.85 | 0.81 |
| $T_{CCS}$ | **0.82** | 0.82 | 0.78 | 0.78 | **0.79** | 0.74 | **0.86** | 0.83 | 0.79 | **0.91** | 0.87 | 0.79 |
| $T_{CUI}$ | **0.82** | 0.81 | 0.79 | 0.78 | **0.79** | 0.76 | 0.85 | 0.82 | 0.74 | **0.91** | 0.87 | 0.79 |

Note: The best performance on each dataset is bold.

**Appendix B. Experimental settings**

In the experiments, we generate representations with the dimensions of 64, 128, and 256 for each patient record. For prediction tasks, the generated representations are used as the input of different classifiers, including logistic regression (LR), random forest (RF), AdaBoost, Neural network (NN), to predict the ICU in-hospital mortalities. We use grid-search to find the best parameters for the classifiers using the MIMIC III - 428 dataset. These classifiers are implemented for all representations generated using the proposed method and baseline methods except the three summary measures (i.e., CCI, ECI, and RSI, which already indicates prediction probabilities) and Med2Vec (which contains a classifier in their research design). The NN uses the Adma optimizer and comprises two fully connected hidden layers, each with 128 neurons, followed by the rectified linear unit (ReLU) activation function and 0.3 dropout regularization, displays best performance in the pilot experiments and is thus chosen as the classifier for all prediction tasks. We grid-search n_neighbors, a parameter in the patient-patient network for the manifold learning algorithms. We evaluate all classifiers' performance through five-fold cross-validation, where the original dataset is randomly split into five equal-sized sub-samples without replacement. The process is repeated in five rounds (i.e., folds). In each round, one single sub-sample is retained as the testing set, and the other four sub-samples are used for classifier training. The classifiers are trained from only the training data of the current round, and the testing data are not seen by the model during the training stage. Please note that the cross-validation is not employed for selecting optimal parameters. All classifiers for different datasets use the same parameters selected using the MIMIC III - 428 dataset. The main reasons for adopting this validation technique are that it achieves a lower bias towards estimating the generalization performance by averaging the individual classifier's estimates (Hastie et al., 2009) and it estimates how the model's performance can be generalized to an independent dataset.